\documentclass[letterpaper, 10 pt, conference]{ieeeconf}  % Comment this line out if you need a4paper

\IEEEoverridecommandlockouts                              % This command is only needed if 
\usepackage{hyperref}

\usepackage{subcaption}
\include{chatbox/chats}

\usepackage{fancyvrb}
\usepackage{fvextra}

\usepackage{xspace}
\newcommand{\methodname}{PerceptTwin\xspace}
\usepackage{algorithm}
\usepackage{rotating} % in preamble
\usepackage{algorithmicx}
\usepackage{algpseudocode}
\usepackage{tikz}
\usepackage{booktabs} 
\usetikzlibrary{fadings}
\usepackage{gradient-text}
\usepackage{array}
\usepackage{adjustbox} % Add this to your preamble
\renewcommand{\baselinestretch}{.985}%9085}
\algrenewcommand\algorithmicindent{0.8em}%
\begin{document}
\bstctlcite{IEEEexample:BSTcontrol} % force et al

\title{\LARGE \bf
\methodname: Semantic Scene Reconstruction for Iterative LLM Planning and Verification\\
\thanks{We acknowledge the support of the Natural Sciences and Engineering Research Council of Canada (NSERC)  [PGS D Scholarships for Charlie Gauthier and Sacha Morin, and funding reference number ALLRP 580895-2022], as well as the support of Denso International.\ (\textit{\small Corresponding author: }\textit{\tt\small  charlie.gauthier@mila.quebec}.)}%
\thanks{$^{1}$Department of Computer Science and Operations Research, Université de Montréal, Montréal, QC, Canada.}
\thanks{$^{2}$Mila - Quebec AI Institute, Montréal, QC, Canada.}
\thanks{$^3$ CIFAR AI Chair.}
%\thanks{*This work was not supported by any organization}% <-this % stops a space
%\thanks{$^{1}$Anonymous Authors,  A Department,
%        An Institution, At An Address, In A Country
        %%}
}

\author{
\href{https://charliegauthier.ca/}{Charlie Gauthier} $^{1,2}$, 
\href{https://sachamorin.github.io/}{Sacha Morin}  $^{1,2}$, 
\href{https://liampaull.ca/}{Liam Paull}  $^{1,2,3}$
}% <-this % stops a space

\makeatletter
\let\@oldmaketitle\@maketitle
\renewcommand{\@maketitle}{\@oldmaketitle
\centering
\includegraphics[width=\textwidth]{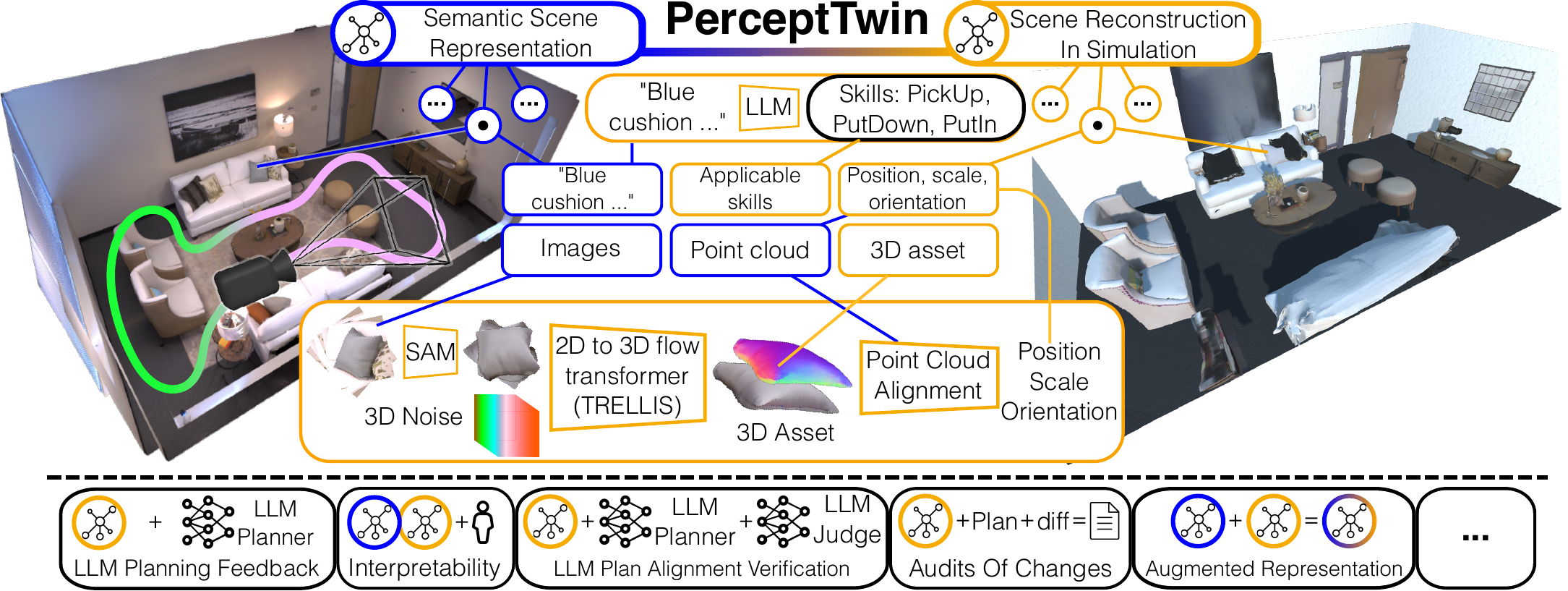}
\captionof{figure}{\label{fig:banner}
State-of-the-art \gradientRGB{robot perception}{0,128,0}{128,0,128} algorithms \cite{conceptgraphs, Werby-RSS-24} build open-vocabulary  \textcolor{blue}{semantic scene representations} that can be used to respond to joint spatial-semantic queries, which is useful for abstract reasoning and planning. \textcolor{orange}{\methodname} consumes such a world representation and generates a corresponding simulation environment. This simulation can then be used for  auditing robot plans, counterfactual analysis, and has the benefit of being more interpretable.  
%\textit{Leftmost scene image credit: \texttt{Room0}, Replica Dataset} \cite{DBLP:journals/corr/abs-1906-05797}.
}
%\vspace{-0.98em}
}
\makeatother

\maketitle
\thispagestyle{empty}
\pagestyle{empty}

%%%%%%%%%%%%%%%%%%%%%%%%%%%%%%%%%%%%%%%%%%%%%%%%%%%%%%%%%%%%%%%%%%%%%%%%%%%%%%%%
\begin{abstract}
Simulation environments are useful for both robot policy learning and planning verification and validation. Traditionally, the process of creating a simulation was onerous. Creating a bespoke simulation environment for each individual environment that a robot would operate in was simply infeasible. In this work, we introduce \methodname, a fully automatic pipeline that constructs interactive simulations directly from semantic scene representations produced by a robot’s perception stack. \methodname combines open-vocabulary object maps with 3D asset generation, affordance prediction, and commonsense condition checking. These interactive simulations can be used to validate and refine plans before they are executed on the robot hardware. Borrowing from the AI alignment literature, we also introduce an LLM judge that verifies plan correctness and alignment with human preferences. 
Experiments show that \methodname feedback allows LLM planners to refine plans, enhance safety, and resist harmful black-box prompting attacks. In our suite of tasks, \methodname improves plan success by an average of $\approx39\%$ for GPT5, GPT5Mini, and GPT5Nano planners. Additionally, \methodname also improves human plan verification by up to $18\%$ on average for plans that fail due to unfilled skill preconditions. Our results demonstrate the potential of open-vocabulary scene simulation from robot perception as a foundation for safer, more reliable robot planning.
\end{abstract}

%%%%%%%%%%%%%%%%%%%%%%%%%%%%%%%%%%%%%%%%%%%%%%%%%%%%%%%%%%%%%%%%%%%%%%%%%%%%%%%%
\section{INTRODUCTION}

Long-held folk wisdom in the robotics community has been that simulations are ``doomed to succeed'' \cite{Brooks1993}, eroding trust that simulated success translates to the real world. Yet, the past decade has revived simulation as a tool for learning (sim2real \cite{andrychowicz2020learning}), where perfection is unnecessary so long as useful information reduces costly robot trials. More recently, particularly in the domain of autonomous driving where data is abundant, we are seeing a return to the verification and validation use case where the simulation itself is generated from real data (real2sim \cite{waymax}).

However, to date this approach has seen limited use in applications other than autonomous driving, such as open-vocabulary scene understanding and planning. In this domain, indoor scene reconstructions are made by simultaneous localization and mapping (SLAM). While recent work \cite{conceptgraphs,Werby-RSS-24} has made SLAM maps more useful for open-vocabulary planning by adding semantic data such as CLIP \cite{clip} embeddings or LLM captions, these representations remain static and passive. 
This is a missed opportunity: most robots operate in everyday scenes with everyday  objects, and we have access to large language models (LLMs) that encode rich commonsense knowledge about how those objects behave. If grounded in an interactive reconstruction of the robot’s environment, an LLM could flag unsafe or incorrect plans since they could actually be rolled out in simulation before attempting execution on real hardware.
%Taking inspiration from the AI alignment and adversarial attack literature \cite{ji2024aialignmentcomprehensivesurvey,zheng2023judging}, we call this LLM a \textit{judge}.

Furthermore, the advent of LLMs has given rise to a new category of planners that reason in natural language \cite{smartllm,liu2024delta}, allowing them to seamlessly integrate with open-vocabulary scene maps. These planner LLMs benefit greatly from obtaining feedback on their initial plans \cite{roco}, suggesting that receiving feedback from a simulation of a scene map could similarly provide substantial benefits.
Also, even ``aligned'' LLMs \cite{ji2024aialignmentcomprehensivesurvey} that should respond with ``safe'' answers can be jailbroken \cite{robey2024jailbreaking}, which is particularly dangerous for robot planning. For example, recent work has shown that  it is  possible for an LLM planner to detonate a bomb next to human simply by reframing the prompt under the guise of an action movie script \cite{robey2024jailbreaking}. %This, too, indicates the need for a process for open-vocabulary scene reconstruction and simulation that would help us provide feedback to LLM planners, or stop them once they've been compromised. 

We address this issue with \methodname, a real2sim pipeline that transforms open-vocabulary 3D scene graphs into interactive simulations suitable for plan verification. 
\methodname (1) finds or generates 3D assets that correspond to objects in the real environment, (2) localizes them using perceived object point clouds, (3) predicts applicable robot-object affordances, (4) is able to test plans inside the simulation and (5) leverages an LLM judge inspired by the AI alignment literature \cite{ji2024aialignmentcomprehensivesurvey} to evaluate  plans. % logic and alignment.%  and refine plans. %This combination improves plan interpretability, correctness, and safety.

Our contributions are: (1) A fully automated real2sim pipeline that consumes open-vocabulary 3D scene graphs and produces interactive simulators, (2) An iterative LLM-based planner that receives feedback from the simulator to refine and align plans, (3) An LLM-based ``plan judge'' that detects unsafe or infeasible plans and suggests corrections, (4) Empirical evidence that \methodname both improves LLM plans and human ability to predict plan success: LLM planning success improves by an average of $39\%$, and human plan prediction accuracy improves by up to $18\%$.
Our code -- from reconstruction to planning to plotting -- is open-sourced at \href{https://percept-twin.github.io}{https://percept-twin.github.io}.

%\section{INTRODUCTION}
\label{sec:introduction}

\definecolor{mypurple}{RGB}{40, 18, 49}
\definecolor{myorange}{RGB}{97, 45, 14}
\definecolor{mydarkred}{RGB}{82, 2, 2}

\section{RELATED WORK}

\begin{comment}
\subsection{Foundation Models}
\methodname relies on foundation models to reason about open-vocabulary objects, as in prior semantic scene representation methods \cite{conceptgraphs,Werby-RSS-24}. CLIP \cite{clip} aligns image and text embeddings for cross-modal comparison, while REMBG \cite{rembg} and SAM \cite{sammodelkirillov2023segany} provide automatic background removal and object segmentation. Large language models (GPT-3.5–5 \cite{openai2024gpt4technicalreport,openai2025gpt5systemcard}) supply commonsense, open-vocabulary reasoning and generate visual captions, and TRELLIS \cite{trellis}, trained partly on Objaverse \cite{objaverse}, produces 3D assets in multiple formats for resource-constrained reconstruction.

\subsection{Semantic Scene Representations}
\label{sec:relwork:conceptgraphs}
3D scene graphs (3DSGs) \cite{armeni20193d} present a unified representation for 3D data and camera views. Objects are represented as nodes with spatial or semantic edges. Recent variants, such as ConceptGraphs \cite{gu2024conceptgraphs} or ``HOVSG'' \cite{Werby-RSS-24}, combine object segmentation (SAM, YOLOv8World \cite{Cheng2024YOLOWorld}) with visual LLM captions, yielding open-vocabulary labels that enhance semantic reasoning. These representations serve as the input to our method which generates scene reconstructions. \methodname assumes an equivalent input: segmented objects with captions and global-frame point clouds.
\end{comment}

\subsection{Open-Vocabulary Semantic Scene Representations}
\label{sec:relwork:conceptgraphs}
3D scene graphs (3DSGs) \cite{armeni20193d} present a unified representation for 3D data and camera views, where objects are nodes linked by spatial or hierarchical edges. Recent extensions, such as ConceptGraphs \cite{conceptgraphs} and HOVSG \cite{Werby-RSS-24}, leverage foundation models to provide open-vocabulary semantics. For example, CLIP \cite{clip} enables cross-modal comparison of images and text, SAM \cite{sammodelkirillov2023segany} enable grounded segmentation, and large language models (GPT5 \cite{openai2025gpt5systemcard}) contribute commonsense reasoning and visual captions. These tools enrich 3DSGs with semantic labels and attributes, forming the input to \methodname: a map containing segmented objects with captions and global-frame point clouds.

\subsection{Scene Generation and Reconstruction}

Recent work on the generation of interactive scenes using foundation models has largely focused on generating scenes given a textual description \cite{sun20253dgeneralistselfimprovingvisionlanguageactionmodels, holodeck, robocasa2024}. Works that do leverage scene graphs only do so as an intermediary representation and still require a textual prompt (and use simplified scene graphs without the richness of real-world 3DSGs) \cite{lin2024instructscene, gao2024graphdreamer}. Other  methods require human input for asset editing or articulation annotation \cite{torne2024rialto}. \textit{\methodname instead directly augments off-the-shelf semantic scene representations, without textual prompts or human intervention.}
Closely related to \methodname are ProcThor \cite{procthor} and Holodeck \cite{holodeck}, both based on AI2Thor \cite{ai2thor}. The former builds novel scenes procedurally and is limited to a closed set of hand-created assets, while the latter generates novel scenes from a text prompt using LLMs and uses CLIP \cite{clip} to semantically search for assets in  Objaverse \cite{objaverse}, a massive dataset of 3D assets. 
%We reuse the CLIP+Objaverse module in the resource-constrained regime (see section~\ref{sec:clip+obj}). 
\textit{Unlike these, \methodname  reconstructs real-world scenes into AI2Thor, an unprecedented capability.}

\begin{comment}

\subsection{Planning}

A longstanding challenge in robotics is developing planners for open domains, particularly in terms of semantic understanding. Traditional planning approaches, such as ``planning domain definition language'' (PDDL) \cite{pddlppdl}, struggle to generalize to open-vocabulary scenes, and are unable to parse the natural language that humans use to specify tasks. 
The recent emergence of LLMs has introduced new possibilities for enhanced understanding of open-vocabulary scenes. Thanks to their ability to retrieve commonsense human knowledge and translate it into machine-understandable code, LLM planners enable a level of context awareness previously deemed impossible \cite{firstllmplanner, liu2024delta, smartllm, roco, openai2024gpt4technicalreport}. 
\textit{In section~\ref{subsec:planning}, we show how \methodname can be used to refine LLM plans, both in terms of safety an plan success.}
\end{comment}

\section{FROM MAP TO SIMULATION}
\label{sec:method}

\begin{algorithm}[b]
\caption{\methodname Reconstruction}
\label{alg:reconstructor}
\begin{algorithmic}[1]
\renewcommand{\algorithmiccomment}[1]{\textcolor{gray}{$\triangleright$ #1}}
\State \textbf{Input:} $M$: semantic scene map built from robot perception where each object $o_i \in O$ has $\texttt{PointCloud}(o_i)$, $\texttt{Images}(o_i)$, and open-vocabulary $\texttt{Description}(o_i)$
\State \textbf{Output:} $\mathcal{S}$: simulated scene containing for each $o_i\in O$ a 3D asset $a_i$, 3D transformation matrix $t_i$, and applicable affordances $s_i$
\State $S\leftarrow$ Initialize empty scene  
\For{$o_i$ in $O$}  \Comment{For each object}
    \State $a_i\leftarrow$ \texttt{AssetFinding}(\texttt{Description}($o_i$), \texttt{Images}($o_i$), \texttt{PointCloud}($o_i$))
    \State $u_i \leftarrow$ \texttt{AssetPlacement}($a_i$, \texttt{PointCloud}($o_i$))
    \State $s_i \leftarrow$ \texttt{PredAffordances}(\texttt{Description}($o_i$))
    \State $\mathcal{S}[i]\leftarrow \langle o_i, a_i, u_i, s_i \rangle$
\EndFor
\end{algorithmic}
\end{algorithm}

\subsection{Problem Statement}

Given a real-world scene ${S}$ and a sequence of sensor observations and control inputs, semantic scene representation methods use SLAM to jointly estimate a robot's trajectory 
and a semantic map of objects
${M} = \{\langle l_i, {g}_i\rangle\}_{i=1}^N$, where $l_i$ contains open-vocabulary natural language information and ${g}_i$ contains geometric information.
Conceptually, we tackle the inverse problem: generating scene $\hat{{S}}$ given map  ${M}$:
\[
\hat{{S}} \sim p({{S}} \mid {M})%\propto \; p({M} \mid {S})\,p({S})
\]

We assume $M$ to be built using modern semantic scene representation methods such as ConceptGraphs \cite{conceptgraphs}, and to contain, for each object: LLM-generated natural language descriptions% (some methods only include semantic embeddings  but \methodname assumes natural language descriptions),
, object-segmented image views (where segmentations are generated by SAM \cite{sammodelkirillov2023segany}), and object-segmented point clouds.
\methodname only has access to the information contained in $M$, i.e., the objects in the scene and their data. For example,  it does not have access to other information such as floor and wall colour if this is not contained within the map representation.

In this section, we describe how \methodname creates an interactive simulation from the input map. As detailed in Alg.~\ref{alg:reconstructor} and Fig.~\ref{fig:banner}, we start by finding or generating adequate 3D assets for each object in $M$, which are then localized and oriented using the object's perceived point cloud. For each object, we use an LLM to predict applicable object affordances from a list of implemented skills (Table~\ref{table:affordances}).

\subsection{3D Assets} \label{subsubsec:af} 
\begin{figure*}
\centering
    \vspace{0.75em}\includegraphics[width=0.8\linewidth]{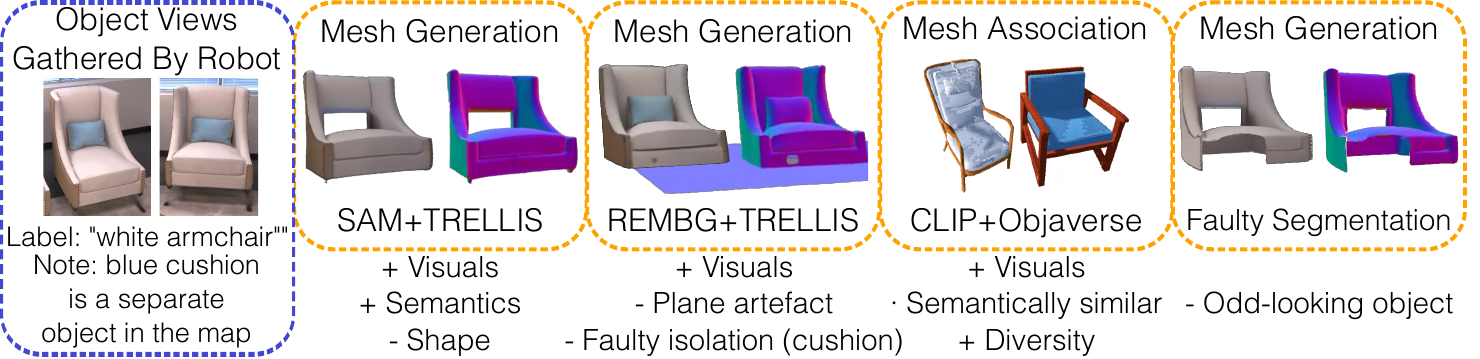}
  \caption{
  Reconstructing \textcolor{blue}{input maps} requires 3D assets, which \textcolor{orange}{\methodname} obtains using TRELLIS or Objaverse. TRELLIS \cite{trellis} originally preprocessed images using REMBG \cite{rembg}. We instead propose to use SAM \cite{sammodelkirillov2023segany}. This improves object isolation and reduces artifacts, improving the semantic closeness of the generated assets with the target object. In both cases, TRELLIS outputs objects with holes when segmentation fails.  CLIP+Objaverse \cite{holodeck} trades visual fidelity for diversity.
  \vspace{-0.5em}
  }
  \label{fig:meshablate}
\end{figure*}

To reconstruct the input $M$, \methodname  requires 3D assets that are semantically and visually aligned with the objects in $M$. 
However, not all downstream tasks will require good visuals or good collision meshes: for instance, symbolic LLM planning cares little for visual similarity, while human interpretability demands adequate visuals.
For this reason, we propose two methods to obtain 3D assets: mesh association and mesh generation. The former is fast and cheap but tends to be less accurate (see Fig.~\ref{fig:reconstructions}, Fig.~\ref{fig:meshablate}), while the latter uses a costly state-of-the-art 2D to 3D transformer. %We compare the two submodules in section~\ref{subsec:afablate}.

\paragraph{Mesh Association} \label{par:ma} \label{sec:clip+obj}
This module, proposed by Holodeck \cite{holodeck}, searches a database of 3D assets by comparing precomputed CLIP \cite{clip} embeddings with a target  query embedding. In Holodeck, the query is obtained from a textual specification suggested by an LLM as part of building a novel scene; in our case, the object already exists in our world and we compute CLIP embeddings from the information contained in the input map $M$ (e.g., natural language descriptions or robot-captured images). 
The top $K$ assets are retained according to the CLIP similarity score\footnote{We use the score formulation and CLIP model used by \cite{holodeck} (OpenCLIP \cite{ilharco_gabriel_2021_5143773} with ViT-L/14 \cite{Radford2021LearningTV} trained on \cite{schuhmann2022laionb}). We use the Objaverse \cite{objaverse} dataset. }:

\vspace{-1em}
\begin{align}
100 \cdot \cos\big( \texttt{CLIP}(\text{real object}),\ \texttt{CLIP}(\text{3D asset}) \big)
\end{align}

Assets with important size deviations from the target point cloud in the input map $M$ are discarded. 

\paragraph{Mesh Generation} \label{par:mg} 
We generate a 3D asset for the target object.
We use TRELLIS \cite{trellis}, a state-of-the-art 2D-to-3D flow transformer that generates 3D assets from images of the target object. The generated assets have visual features that are greatly inspired by the real appearance of the objects, as shown in Fig.~\ref{fig:banner}. This helps match the visuo-semantics of the real objects, as shown in Fig.~\ref{fig:reconstructions}.

\subsection{3D Asset Placement} \label{subsubsec:mf}
To localize objects, we rely on the segmented global-frame point clouds from the input $M$. %This assumption is not restrictive, as \cite{conceptgraphs} can produce single-frame, globally referenced points. 
Estimating object orientation is more challenging: the target point clouds are noisy sensor measurements, while the 3D assets to be aligned are sourced from Objaverse \cite{objaverse} or generated by TRELLIS \cite{trellis}. We found that standard alignment techniques such as iterative closest point (ICP) \cite{924423} or global registration \cite{RANSAC} frequently fail under these conditions. Empirically, we achieve more reliable results with a constrained variant of ICP that assumes horizontal alignment, disables shear transformations, and scales  while preserving  aspect ratio.
The bounding boxes of the point clouds are also used to compute relational edges such as \texttt{mug isOnTopOf table} in the case where the edges are lacking from $M$.

\subsection{Robot-Object Interactions} \label{subsec:affordancepred} 
In robotic symbolic planning, it is common to define multiple supported \textit{skills} such as \texttt{OpenObject}, \texttt{CutObject}, \texttt{PickupObject}, etc. 
We report the list of skills that we implement in \methodname in Table~\ref{table:affordances}.
We motivate this selection by basing it on a subset of the skills listed by the recent LLM planning approach SMART-LLM \cite{smartllm}.

%If required, \methodname's implementation readily accommodates new affordances by compositionally applying preconditions.

In the open-vocabulary domain, it is not possible to know which skills apply to which objects \textit{a priori}. We rely on an LLM to selects affordances from Table~\ref{table:affordances}, given information about the target robot (number of arms, height, etc.)
as well as the target object (label and caption from $M$).
The LLM also tags appropriate objects as \texttt{slicing implement} to support \texttt{SliceObject}, such as knives or utility blades.

The question remains of how these skills interact with each other. Under the assumption that the final deployment platform will be a single-arm robot, we implement simple commonsense preconditions for each skill. We ensure that the robot can only manipulate one object at once, that a \texttt{pickupable} object must be in hand for it to be \texttt{PutDownObj} on a \texttt{canReceive} object, etc.

\begin{table}[b]
\vspace{-1em}
\small
\captionof{table}{\label{table:affordances} Implemented Robot-Object Interactions}
\centering
\begin{tabular}{p{2.0cm}p{1.9cm}p{2.8cm}}
\toprule
{\footnotesize \textbf{Affordance} $\in P$ } & {\footnotesize \textbf{Skill} $\in A$} & {\footnotesize  \textbf{Preconditions} $\in P$}\\

\midrule
\texttt{pickupable}               & PickupObject     & Hand free, is nearby \\
\texttt{canReceive}                  & PutObject        & Holds \texttt{pickupable} object,  is nearby \\
\texttt{canContain}              & PutObjectIn      & Holds \texttt{pickupable} object, is nearby, is open \\
\texttt{sliceable}                & SliceObject      & Holds slicing implement, is nearby \\

N/A          & GoToObject       & Path is navigable \\
\texttt{breakable}                  & BreakObject      & Is nearby, unbroken, hand free \\
\texttt{canTurnOnOff}               & ActivateObject   & Is nearby, hand free \\
\texttt{canTurnOnOff}           & DeactivateObject & Is nearby, hand free \\
\texttt{openable}                 & OpenObject       & Is nearby, hand free \\
\texttt{closeable}                & CloseObject      & Is nearby, hand free \\
%\texttt{cleanable}                & CleanObject      & Is nearby, cleanable \\
\bottomrule
\end{tabular}
\end{table}

%\vspace{0.2cm}

In summary, the reconstruction process transforms an input scene map $\mathcal{M}$ into a simulated scene $\mathcal{S}$ which is comprised of a set of assets, which each have their own 
%set of the methods that define the 
skills with which the robot agent can interact with them.   

\section{PLAN VERIFICATION}
\label{sec:methodplan}

The simulation generated by \methodname enables a feedback-enabled planning pipeline: the robot scans a scene, builds a scene map, \methodname builds a simulation, an LLM planner proposes a plan given a high-level task, and iterative feedback in simulation allows the LLM to refine the plan before deployment. In this section, we detail the open-vocabulary plan verification  capabilities of our method.

\subsection{Problem Statement}
\label{subsec:planning_formalism}

We adopt the LLM planning formalism from ProgPrompt \cite{progprompt}. 
A planning task is a tuple 
$\langle O, P, A, T, I, G, t \rangle$, where  
$O$ is the set of objects,  
$P$ is the set of object properties (object states such as \texttt{isSliced}, relational edges such as \texttt{distanceToOtherObj}, affordances from Table~\ref{table:affordances}),  
$A$ is the set of all available actions (all skills in  Table~\ref{table:affordances}),  
$s \in S$ is an assignment of object properties,  
$T : S \times A \rightarrow S$ is the transition model,
$I$ and $G$ denote sets of viable initial and goal states, and $t$ is the planning horizon. The planner only observes a natural language goal description $\mathcal{G}$. 

A plan $P = \langle a_0, \dots, a_{n-1} \rangle$ is a sequence of actions that drives an initial $s_0 \in I$ to a  state $s_n$
; but $P$ might terminate early if an action's preconditions $\texttt{pre}(a)$ are not met.

%\subsection{Precondition Feedback}

\subsection{LLM Planning And Judge Feedback}
\label{subsec:llmjudge}

In recent years, LLM planning formalisms have emerged that enable unprecedented open-vocabulary semantic planning capabilities in robotics \cite{liu2024delta,smartllm}. To perform planning, LLM planners assume access to a textual representation of the world.  \methodname enables full environment serialization into a textual format, as shown in Fig.~\ref{fig:state-diff}.

\methodname can help refine LLM-generated plans with an initial layer of feedback by providing  hardcoded error messages when skills violate their preconditions (Table~\ref{table:affordances}). But
verifying the validity of robot plans in open-vocabulary environments is challenging. The space of possible objects and tasks is vast, and ensuring that a generated plan not only executes but also achieves the correct outcome can require commonsense reasoning beyond hardcoded preconditions. To address this, we augment \methodname  with a trusted LLM. Taking inspiration from the AI alignment and jailbreaking literature, we call this LLM the ``judge'' \cite{zheng2023judging,ji2024aialignmentcomprehensivesurvey}. The judge receives the user-provided planning prompt, but is otherwise isolated from the LLM planner (it only observes the results of the planner's actions); it is compatible with any planner, so long as it allows a feedback mechanism.

\begin{figure*}[htb]
  \centering
  \setlength{\tabcolsep}{0pt} % adjust spacing between columns
  \renewcommand{\arraystretch}{0.0} % adjust spacing between rows
\vspace{0.5em}
  \begin{tabular}{>{\centering\arraybackslash}l c c c c c} % first column: fixed width, vertical centering
  % Row 1: Yard
  \raisebox{0.5\height}{\begin{sideways}Backyard\end{sideways}} &
  \includegraphics[width=0.18\textwidth]{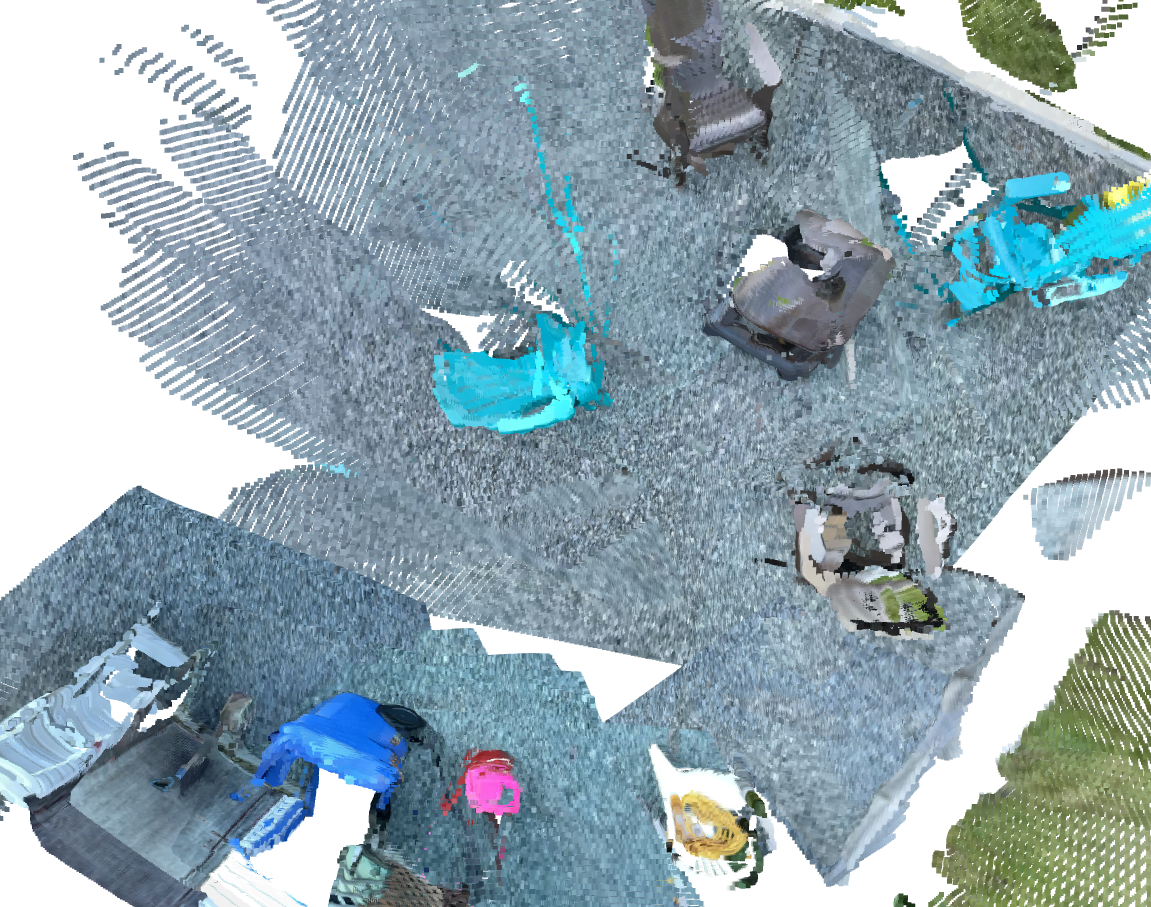} &
  \includegraphics[width=0.18\textwidth]{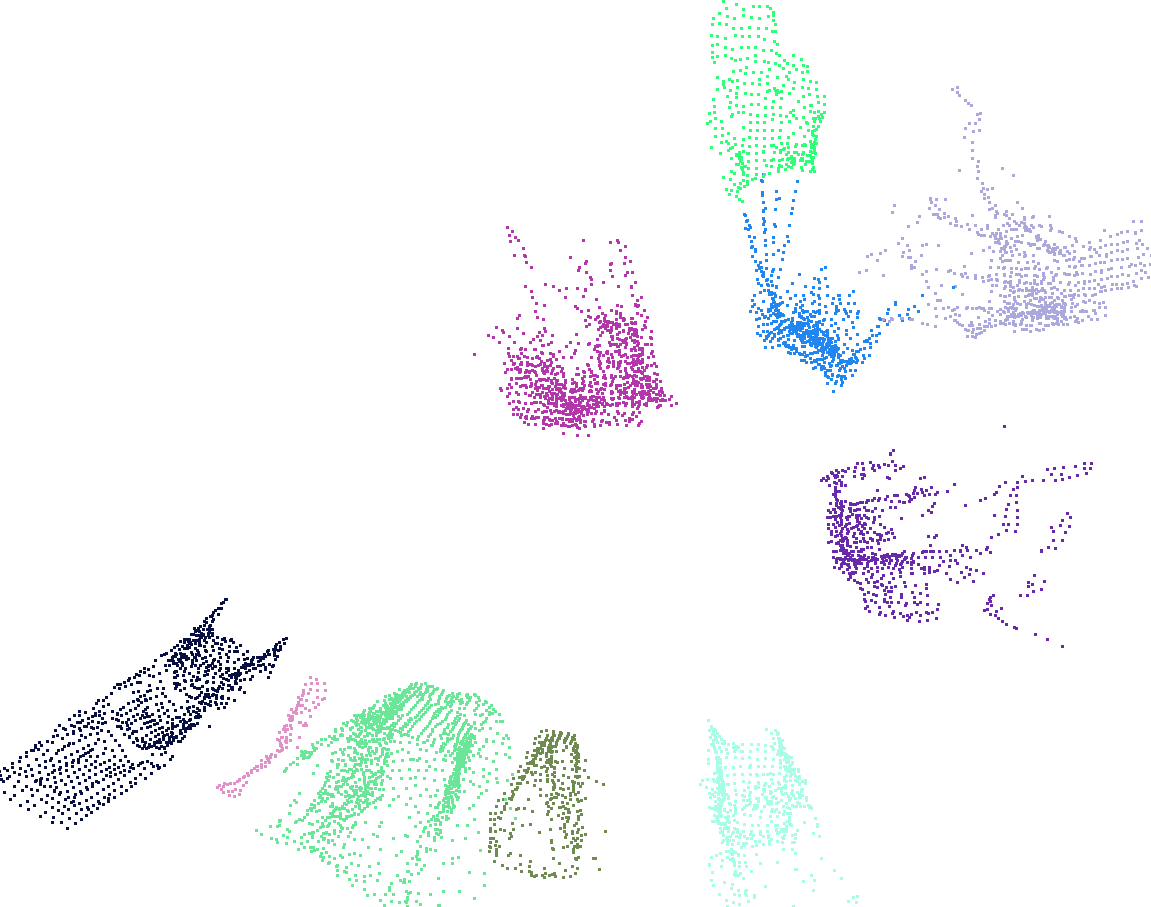} &
  \includegraphics[width=0.18\textwidth]{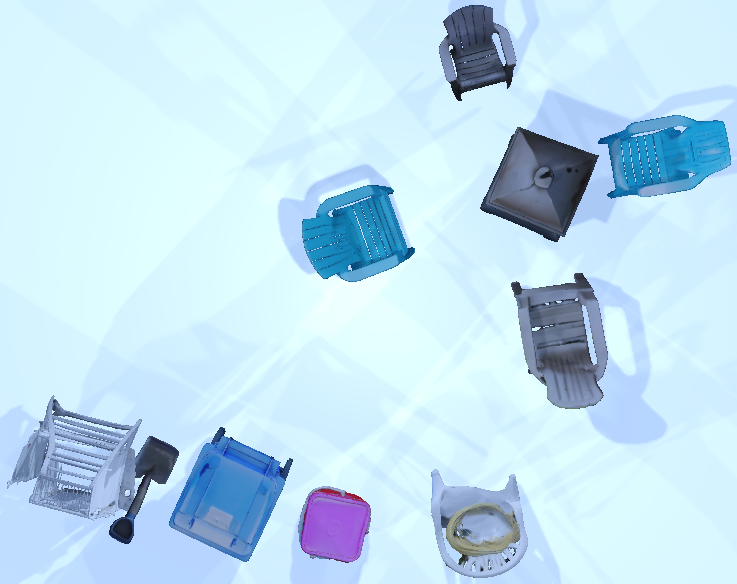} &
  \includegraphics[width=0.18\textwidth]{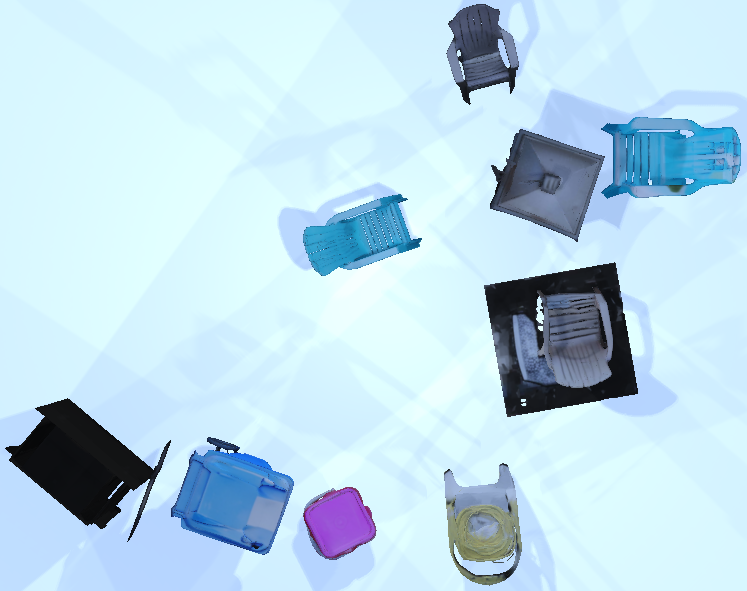} &
  \includegraphics[width=0.18\textwidth]{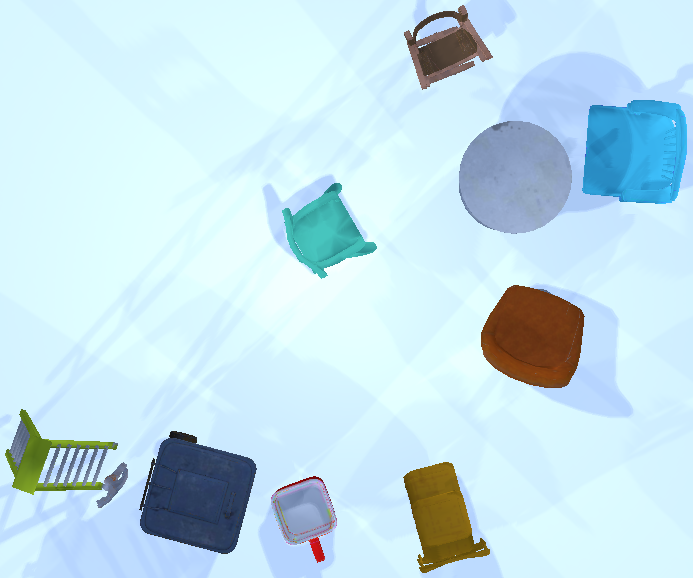} \\

  % Row 2: Cones
  \raisebox{1.5\height}{\begin{sideways}Cones\end{sideways}} &
  \includegraphics[width=0.2\textwidth]{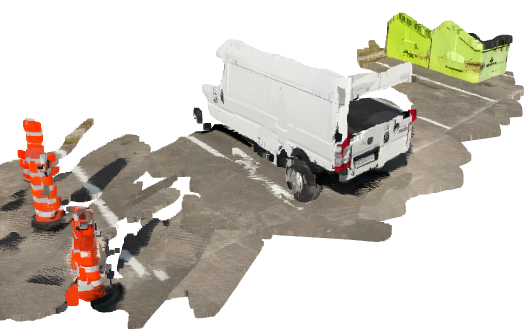} &
  \includegraphics[width=0.2\textwidth]{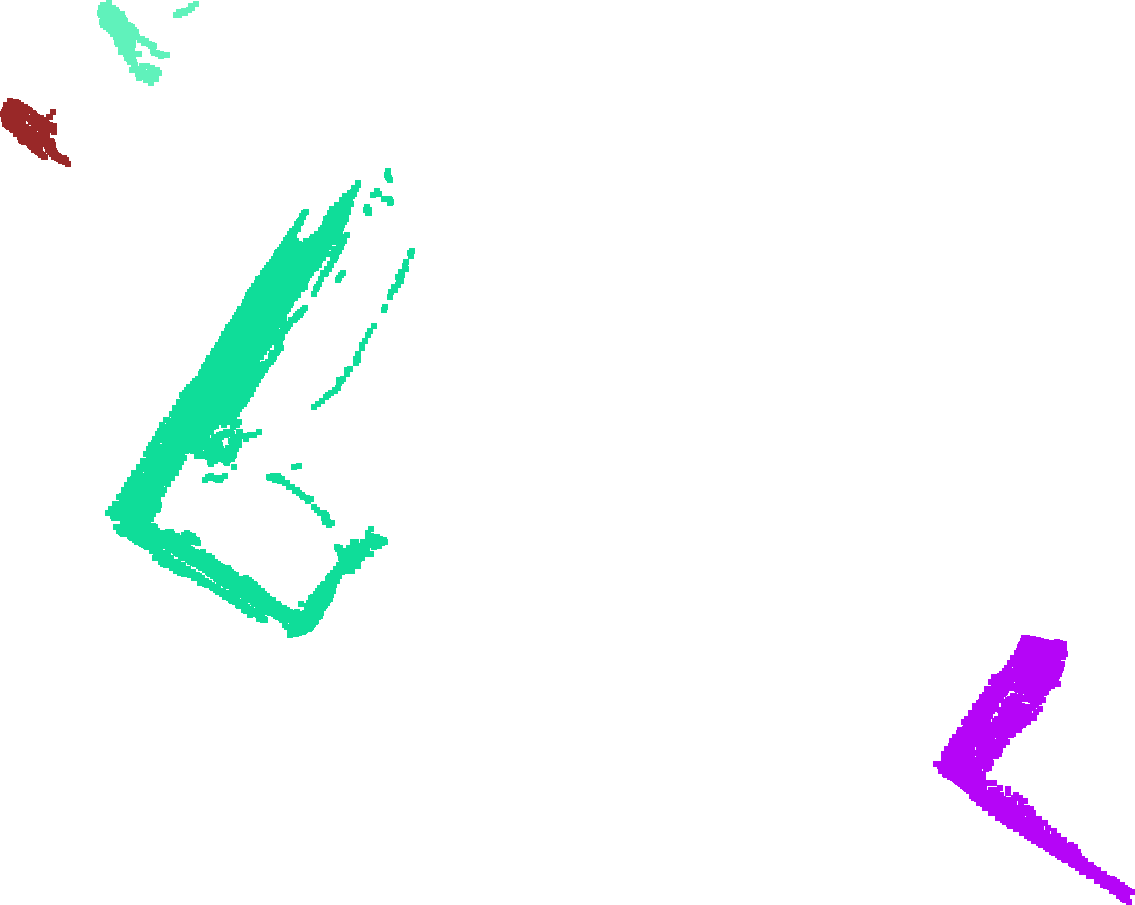} &
  \includegraphics[width=0.18\textwidth]{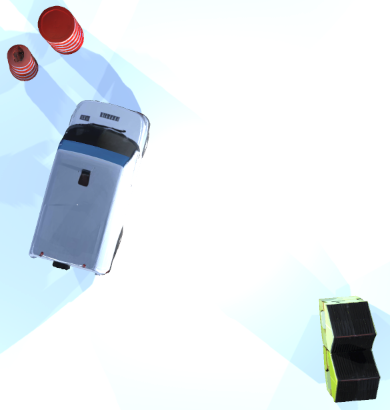} &
  \includegraphics[width=0.18\textwidth]{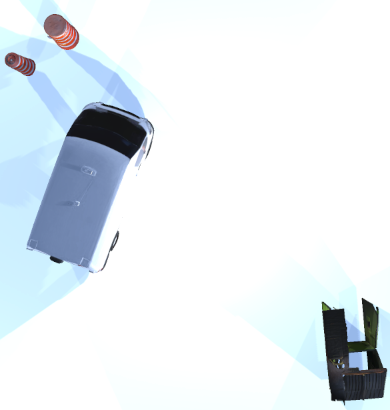} &
  \includegraphics[width=0.18\textwidth]{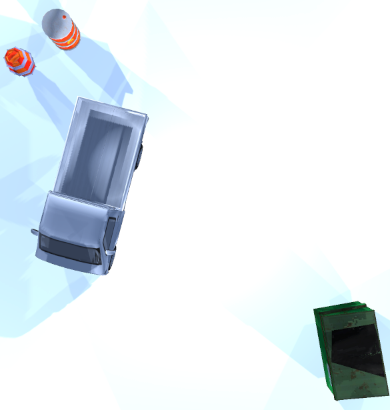} \\

  % Row 3: Blocks
  \raisebox{1\height}{\begin{sideways}Blocks\end{sideways}} &
  \includegraphics[width=0.09\textwidth]{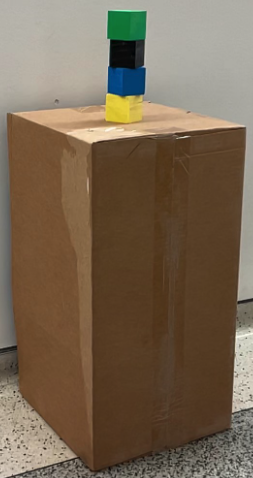} &
  \includegraphics[width=0.15\textwidth]{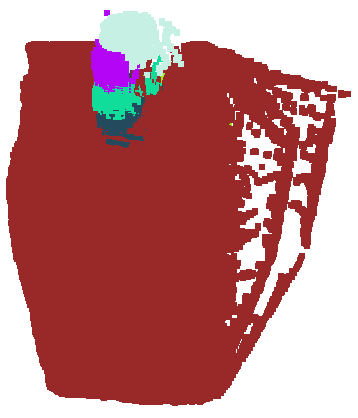} &
  \includegraphics[width=0.12\textwidth]{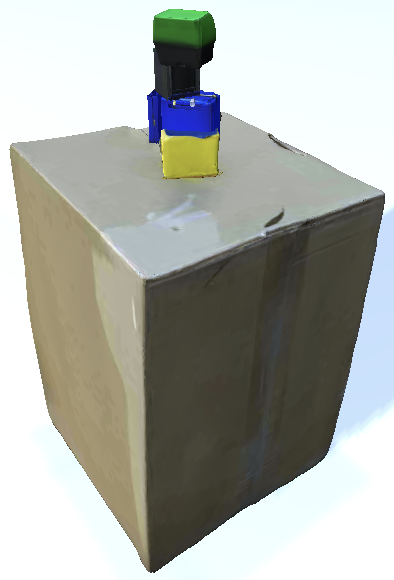} &
  \includegraphics[width=0.12\textwidth]{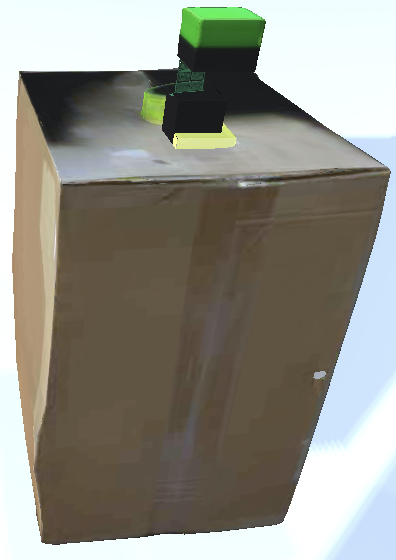} &
  \includegraphics[width=0.11\textwidth]{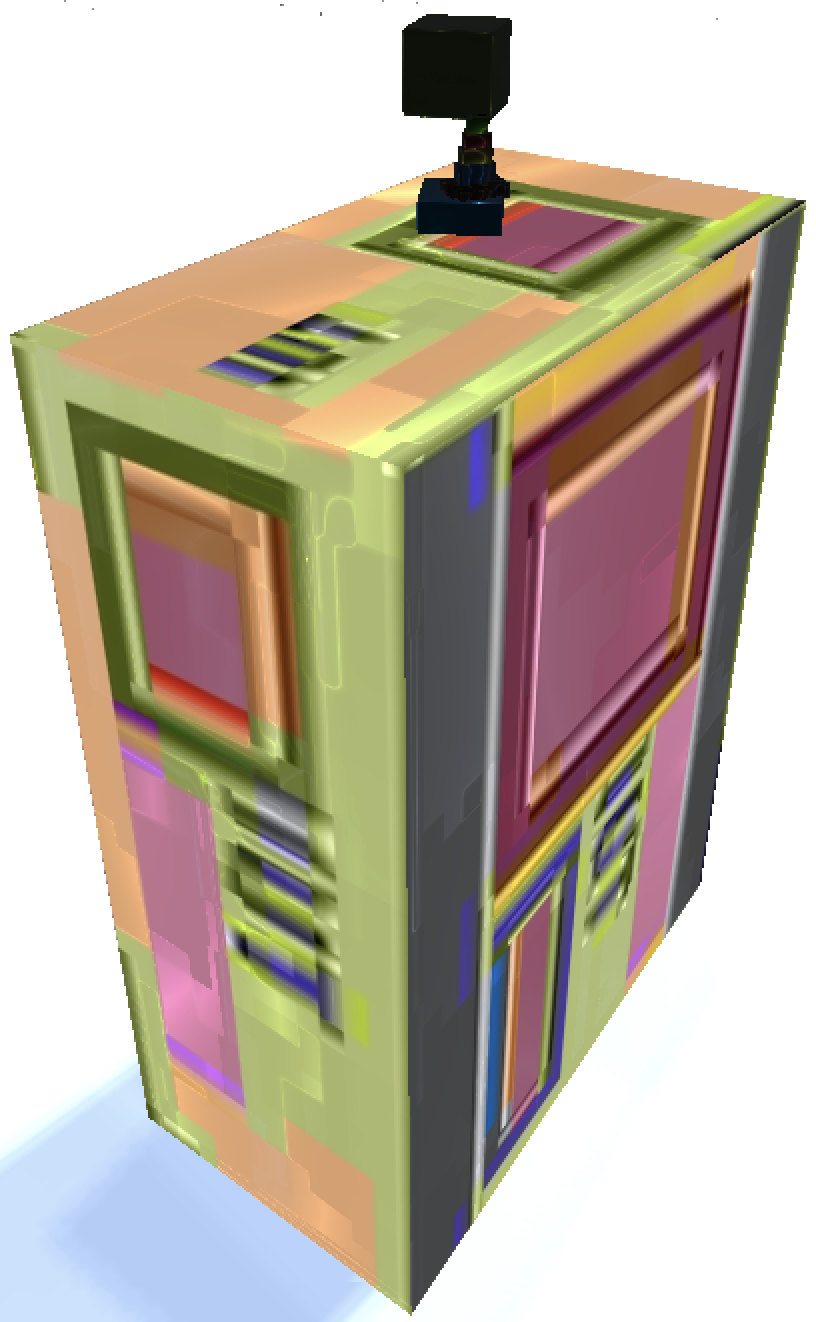} \\

  % Row 4: Packing Veggies
  \raisebox{0.25\height}{\begin{sideways}Veggies\end{sideways}} &
  \includegraphics[width=0.19\textwidth]{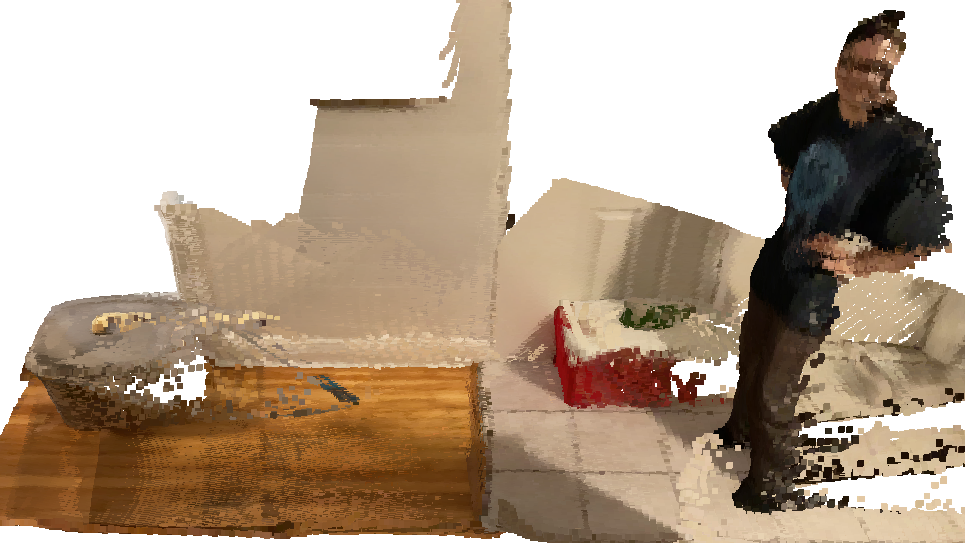} &
  \includegraphics[width=0.19\textwidth]{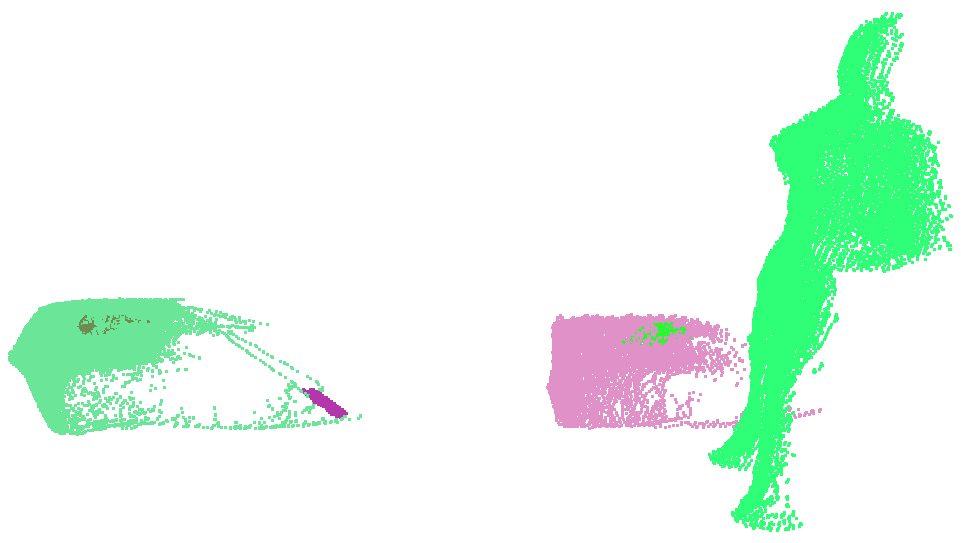} &
  \includegraphics[width=0.18\textwidth]{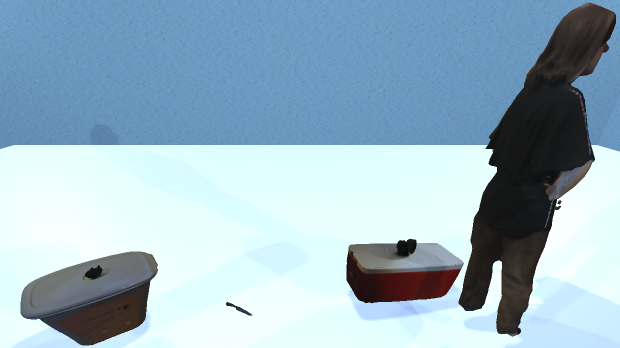} &
  \includegraphics[width=0.18\textwidth]{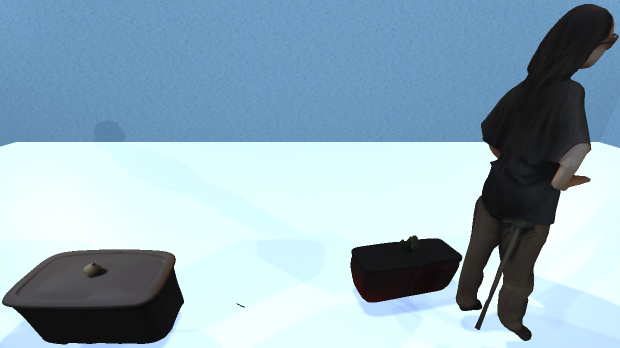} &
  \includegraphics[width=0.18\textwidth]{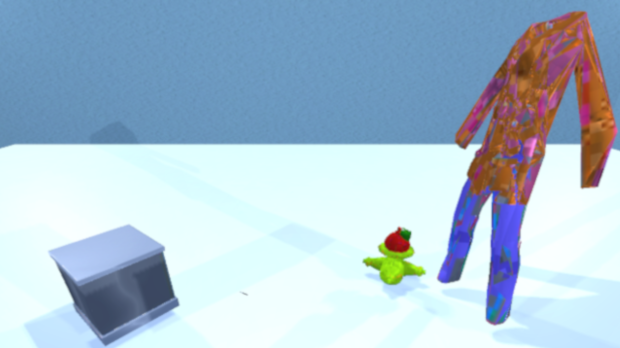} \\

    &\vspace{1em}&&&& \\
  % Column captions
  &
  Photo or raw point cloud & \textcolor{blue}{ConceptGraph \cite{conceptgraphs}} & \textcolor{orange}{SAM+TRELLIS (ours)} & \textcolor{orange}{REMBG+TRELLIS \cite{trellis}} & \textcolor{orange}{CLIP+Objaverse \cite{holodeck}} \\
\end{tabular}

  \caption{\label{fig:reconstructions}\methodname reconstructs diverse input maps, spanning large objects and both indoor and outdoor scenes (see also Fig.~\ref{fig:banner}). Although photorealism is unattainable due to the  limited \textcolor{blue}{input data} (e.g., missing floor color, restricted viewpoints), our SAM+TRELLIS pipeline yields more accurate assets than CLIP+Objaverse (e.g., pickup truck for van, multicolored box for cardboard box) and REMBG+TRELLIS (e.g., black patches on box, black plane on Adirondack chair). Scenes: \textit{Backyard:}Adirondack chairs around a metal fireplace; ladder, shovel, recycling bin, pink bucket, white plastic chair. \textit{Cones:} two tall cones, white minivan, double dumpster. \textit{Blocks:} cardboard box, “green on black on blue on yellow” block tower. \textit{Veggies:} onion on gray bin, knife, bell pepper on red-and-white cooler, human.
  }
  \vspace{-1em}
  
\end{figure*}

The role of the judge is twofold. First, it verifies \textit{logical correctness}, in that the sequence of skills is consistent with the task. For instance, in the \textit{Veggies} scene shown in Fig.~\ref{fig:reconstructions}, the task ``prep the veggies'' should result in slicing both vegetables. This requires natural language reasoning to  relate the task  to the simulation's state after executing the plan. 

Second, the judge verifies that the plan is \textit{aligned} with human preferences. 
This requires more than just validating plan logic and preconditions. 
While alignment can be encouraged during LLM training~\cite{ji2024aialignmentcomprehensivesurvey}, LLMs remain vulnerable to jailbreaks~\cite{robey2024jailbreaking}. 
For example, an adversary could 
%use a jailbreak to 
fool an LLM planner into using a bomb to harm humans under the guise of a movie script~\cite{robey2024jailbreaking}. 
Such a \textit{misaligned} plan would technically satisfy the skill preconditions and stated goal. %By grounding the judge in 
%\methodname allows us to reason about plan alignment prior to deployment.

To perform verification, we ground the judge on \methodname's simulation. The judge observes the state both before and after plan execution. To minimize token usage, we encode the states textually as key-value pairs and compute their UNIX \texttt{diff}. %, which compactly reports the states and the changes between them. 
An example is shown in Fig.~\ref{fig:state-diff}. This representation contains object-object distances, relational edges, object properties, etc., allowing the judge to focus on the salient changes introduced by the plan. % and evaluate whether they are correct and aligned. 
Again, borrowing from the adversarial attack literature \cite{zheng2023judging,ji2024aialignmentcomprehensivesurvey}, we assume the judge  (and the object-labelling LLM from the input scene representation method) to be isolated from user tampering.% We task the judge with ensuring alignment of the potentially-compromised planner.

In the case of an adversarial prompt as demonstrated in \cite{robey2024jailbreaking}, to verify alignment without compromising the judge, the judge can be applied step-by-step, after each skill execution instead of after plan execution (though this is costly).
Upon invocation, the judge can send one of three types of messages: \texttt{Unsafe <reason>}, \texttt{Incorrect <reason>}, and \texttt{Correct <reason>}. We report sample messages and the judge's reasoning in section~\ref{sec:planresults}.

\begin{figure}[t]
  \centering
  \begin{minipage}{0.9\linewidth}
    \begin{Verbatim}[commandchars=&QW, breaklines=true]
diff  ./state_before.json ./state_after_&fvtextcolorQredWQPickUpObject ("mug")W.json
    ..., "mug": {
    - "isHeld": false,
    + "isHeld": true,
    "object_distances": {
      - "robot": 0.75 meters,
      + "robot": 0.25 meters, 
      ...
    }, "object_isOnTopOf": [
      - "table"
    ],...
    \end{Verbatim}
  \end{minipage}
  \caption{\methodname computes \texttt{diffs} between scene states for succinct audit reports of plans or individual skills.\vspace{-10px
  }}
  \label{fig:state-diff}
\end{figure}

\section{RESULTS}

%\subsection{Visuals}

%\section{RECONSTRUCTION RESULTS}
\label{sec:reconstructionresults}

\label{sec:downstreamtaskresults}

\begin{figure*}
\centering
\vspace{0.5em}
\includegraphics[width=0.24\textwidth]{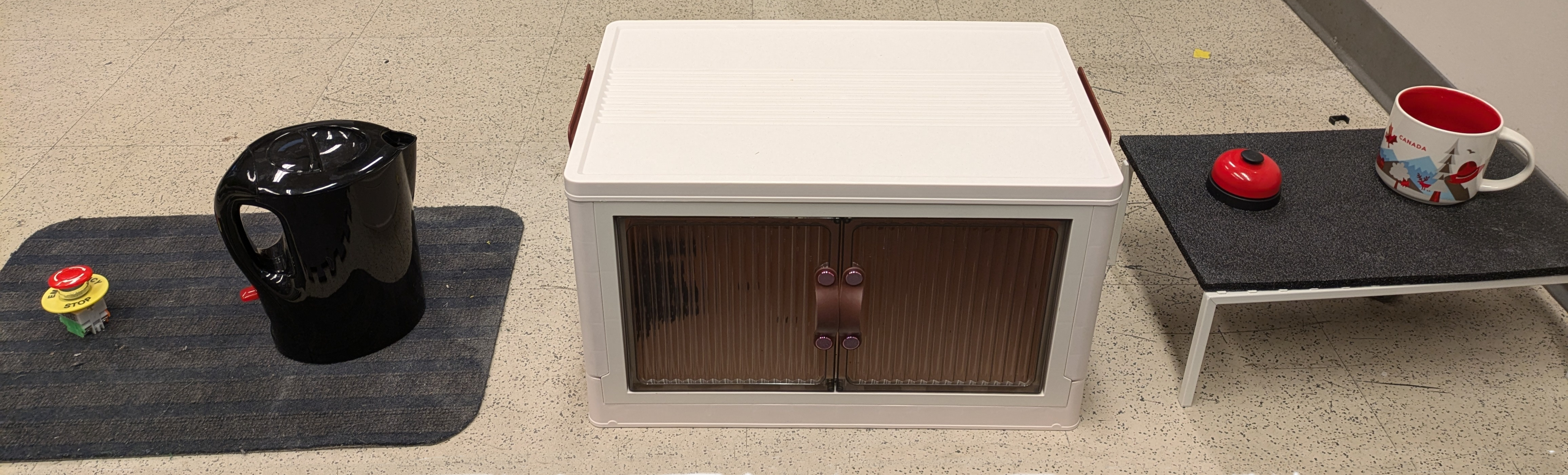} \hspace{-0.7em}
\includegraphics[width=0.21\textwidth]{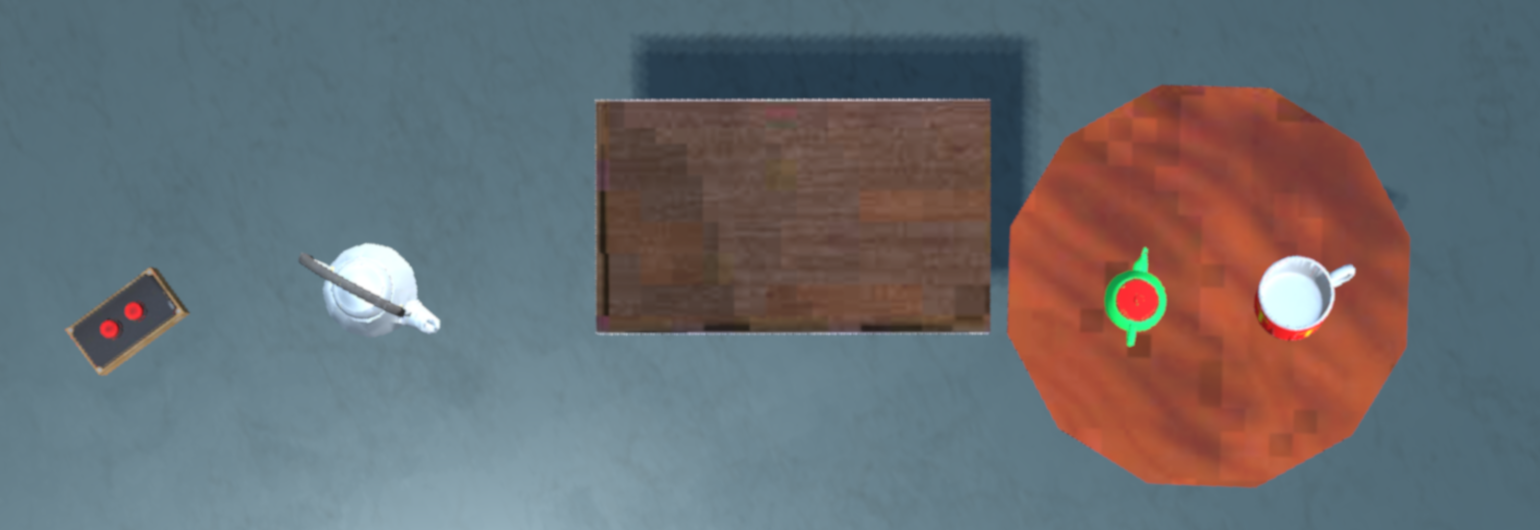} \hspace{-0.7em}
\includegraphics[width=0.55\textwidth]{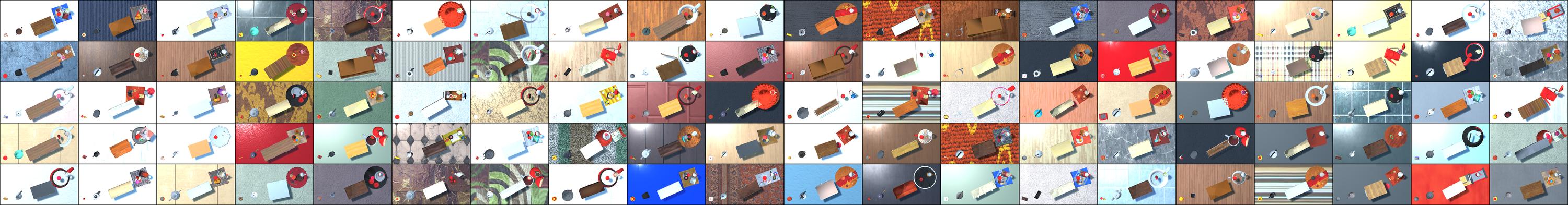}
\caption{
\label{fig:scenegen}
\textit{An emergency stop button, a black kettle, a white container, a small table, a desk bell, a mug.}
Diverse CLIP+Objaverse \cite{clip,objaverse} reconsturctions, generated without human supervision from a ConceptGraph \cite{conceptgraphs} collected using a LoCoBot  and an Intel Realsense. The floor colourings were obtained at random from AI2Thor's \cite{ai2thor} large selection of floors.
\vspace{-1.5em}
}
\end{figure*}

In this section, we provide evidence for our claim a simulated reconstruction of a semantic scene map can be broadly useful for robotics. We begin with reconstruction results. Then, we move onto downstream tasks.
Beyond familiar applications like dataset generation (shown in Fig.~\ref{fig:scenegen}), we argue that \methodname's value lies in three key areas: enhancing human interpretability,  plan alignment verification, and iterative feedback for LLM-based planners.
In all experiments, the input scene map is built using  ConceptGraph \cite{conceptgraphs}.

\subsection{Visuals}
\label{subsec:afablate}

\begin{figure}
    \centering
    \includegraphics[width=0.5\linewidth]{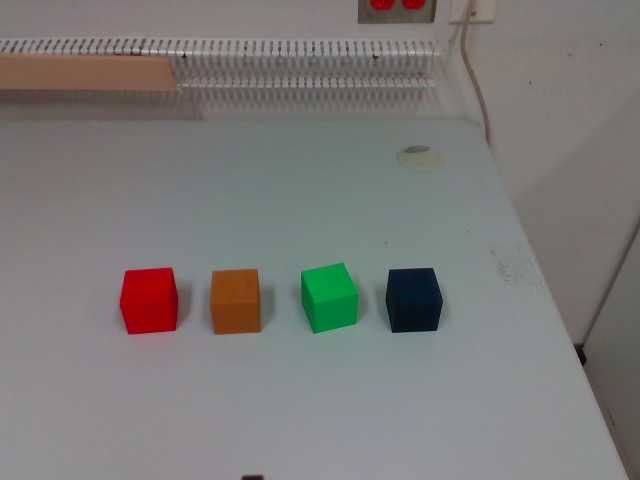}
    \hspace{-10px}
    \includegraphics[width=0.5\linewidth]{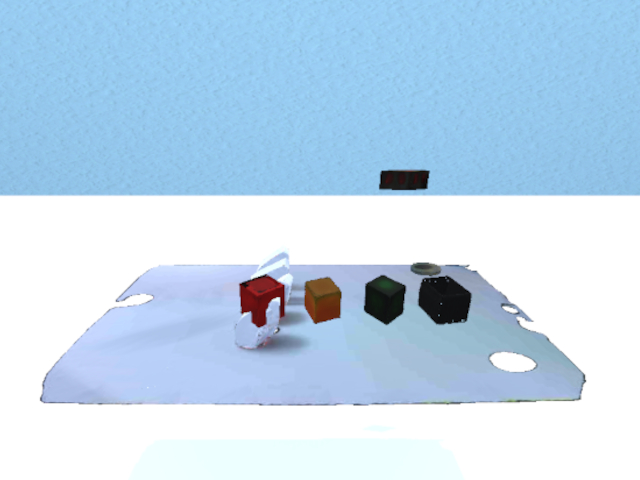}
  \caption{\label{fig:singleframe}The scenes in Fig.~\ref{fig:reconstructions} were generated from longer SLAM trajectories, whereas here, the input ConceptGraph \cite{conceptgraphs} was built from a single frame captured by an Intel Realsense\texttrademark\ mounted on a robot arm. This scene highlights another useful aspect of \methodname: the scene is quite representative of the input map, down to the raised bumps and the holes in the TRELLIS-generated \cite{trellis} table surface that reflect the masks left behind by SAM \cite{sammodelkirillov2023segany} segmentation. \vspace{-1.5em}}
\end{figure}

Fig.~\ref{fig:meshablate} illustrates that TRELLIS \cite{trellis} often better captures visuo-semantics than the CLIP+Objaverse submodule from Holodeck \cite{holodeck}.
However, the original preprocessing for TRELLIS used REMBG \cite{rembg} to segment objects of interest in input images; we find this to be unsuitable.
%, as REMBG cannot be conditioned to isolate target objects in multi-object images. We also find that REMBG+TRELLIS often produces artifacts such as planes or  black patches (Fig.~\ref{fig:reconstructions}). 
Given that we assume access to an input $M$ with SAM-segmented object views \cite{sammodelkirillov2023segany}, we replace REMBG with these object-specific segments. Reconstructions in Fig.~\ref{fig:reconstructions} suggest that SAM+TRELLIS tends to more accurately reflect target objects.  Fig.~\ref{fig:singleframe} shows that we  can also reconstruct input maps made from a single robot-perceived image.

\textit{Time and hardware requirements:} On a standard workstation, processing Fig.~\ref{fig:banner} ($\approx 30$ objects) using the CLIP+Objaverse \cite{holodeck} module took $\approx 5$  minutes and $\approx 8GB$ RAM. No GPU was necessary.
The TRELLIS \cite{trellis} module  requires an NVIDIA GPU of the AMPERE architecture or newer with at least 16 GB of VRAM \cite{trellis}; we used an NVIDIA L40S. Processing Fig.~\ref{fig:banner} took {$\approx 1$} hour.

\begin{comment}

\subsection{Affordance Prediction}

We design a retrieval task to assess the impact of LLM performance on skill-object prediction. The model receives a prompt containing an object label and a set containing one ground-truth and nine randomly sampled distractors. To decouple prediction ability from the specific skills that we chose to implement (and to run a large-scale evaluation), we draw verbs and objects from the HICO-DET dataset \cite{zhang2021scg}. We remove references to the target robot from the prediction prompt to match HICO-DET's human focus. An example task is ``water bottle --- crush, fill, switch on, twist, run, ...'' 
A trial is counted as successful if we find the ground-truth action in the list output by the model. We repeat the task 50 times per GPT \cite{openai2025gpt5systemcard} version (2000 total API calls) using a fixed task seed. GPT-3.5, GPT-4, GPT-4.1, and GPT-5 respectively achieved accuracies of $0.66 \pm 0.22$, $0.88 \pm 0.10$, $0.87 \pm 0.14$, and $0.80 \pm 0.14$. Based on these results, we employ GPT-4 for the affordance prediction submodule.

\end{comment}

\begin{comment}

\begin{table}[b]
\centering
\small
\caption{HICO-DET \cite{zhang2021scg} Action Verb Retrieval Accuracy}
\label{table:hicodet}
\begin{tabular}{lcccc}
\toprule
GPT-3.5 & GPT-4 & GPT-4.1 & GPT-5  \\
\midrule
  $0.66 \pm 0.22$ & $0.88 \pm 0.10$ & $0.87 \pm 0.14$ &  $0.8 \pm 0.14$  \\
\bottomrule
\end{tabular}
\end{table}\end{comment}

\subsection{Human Interpretability}
\label{sec:human-interpretability}

We evaluate whether \methodname (using the mesh generation strategy) improves human interpretability of robot plans through a user study. Under A/B testing, participants predicted plan success using: (A) a \textit{baseline} video of the input map’s point cloud from multiple views, or (B) a plan execution video in \methodname. In both cases a representative image of the scene was also provided. %\todo{We used the \textit{Mesh Generation} strategy.}

Survey questions fell into one of two categories: \textit{logic}, where all skills executed but the final state could be correct or incorrect, and \textit{consistency}, where execution failed due to violated preconditions or other structural mistakes. The former probes reasoning about outcomes, while the latter assesses detection of process-level failures. In the latter, \methodname can display explicit error messages, so we expect strong positive effect on participant accuracy.  

%Survey questions fell into one of two categories: 
%\textit{logic}, where all affordances executed but the final state could be either correct or incorrect, and \textit{consistency}, where execution failed due to violated preconditions, ordering mistakes, or other structural errors. The former probes participants’ ability to reason about outcomes,
% while the latter assesses their ability to detect process-level failures.
% In the latter, \methodname can explicitly display the reason for a failure as an error message; the interesting aspect of the experiment is the scale of the improvement.

\begin{figure}
\centering
    \adjustbox{valign=c}{\includegraphics[width=0.75\linewidth]{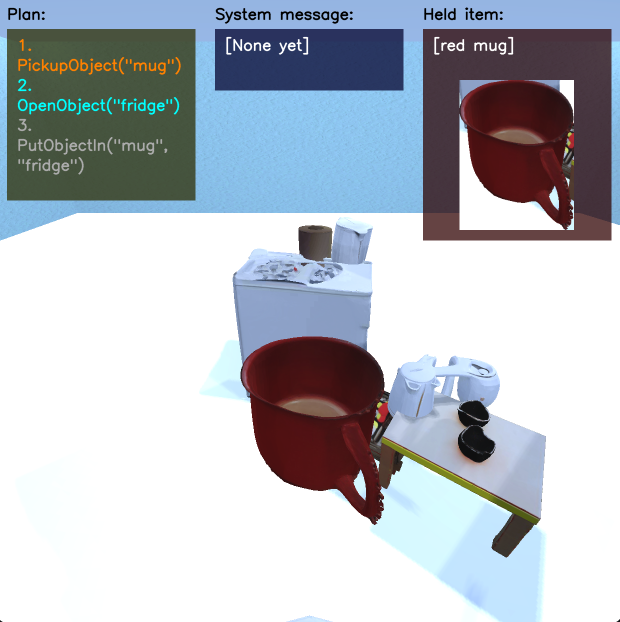}}
    \hspace{-20px}
    \adjustbox{valign=c}{\includegraphics[width=0.30\linewidth]{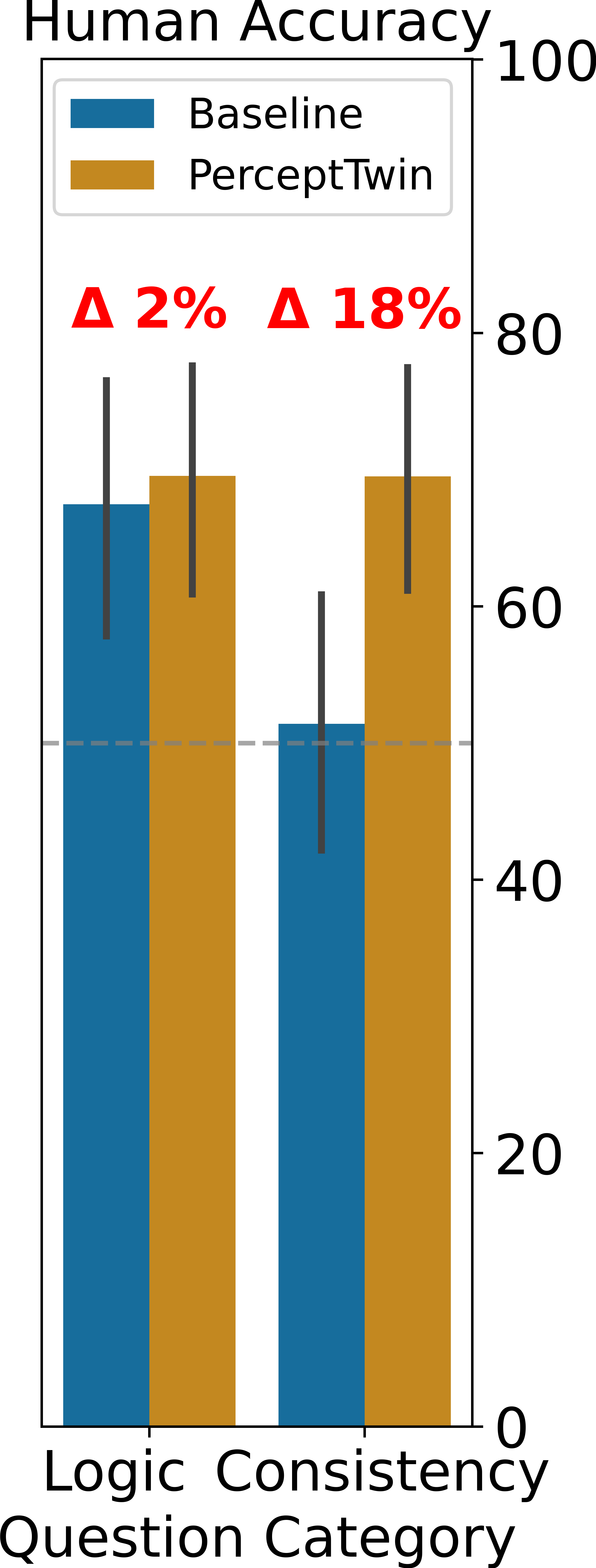}}
  \caption{ \textit{A brown paper towel roll and water pitcher on a fridge. A mug and a kettle on a  table.} Participants ($N=93$) were asked to predict plan success or failure given a video of the input scene map's point cloud (baseline) or  a simulated video of the plan (\methodname); one example \methodname video frame is shown on the left. On the right, we average over all tasks and scenes.
Statistical tests revealed significant improvements in \textit{consistency} questions. 
  %that \methodname significantly improved participant accuracy for plans with precondition failures.
  \vspace{-1.5em}}
  \label{fig:humanviewing}
  
\end{figure}

We designed five scenarios: In \textit{Blocks} (Fig.~\ref{fig:reconstructions}), we use plans that reorder the tower: two \textit{logic} questions (one success, one failure) and one \textit{consistency} question  (all blocks were \texttt{Picked} before being \texttt{PutDown}, infeasible for  single-arm robots). In \textit{Cones} (Fig.~\ref{fig:reconstructions}), one \textit{consistency} question (the plan attempted to place a cone atop a dumpster and was flagged as infeasible due to the specified robot being too short). In \textit{Kitchen} (Fig.~\ref{fig:humanviewing}), a \textit{consistency} failure arose when the robot tried \texttt{OpenObj} while already holding an object.

The dependent variable is participant accuracy in predicting plan success/failure; the independent variables are question category and visualization. We conducted a two-way repeated-measures $t$-test with Holm's correction \cite{29def780-e117-38f0-8afb-edf384af3fad} (we assume normality due to large $N$). While both categories improved under \methodname, statistical significance was observed only for \textit{consistency}.  
We hypothesize that this higher \textit{consistency} increase reflects human bias in interpreting natural-language skill names. 
 For instance, even though participants were aware that the target robot had a single arm,  the \texttt{PickUp} skill means something different for most humans (dexterous, two-armed) than for a single-armed robot: for the ``place the mug in the fridge'' task shown in Fig.~\ref{fig:humanviewing}, most humans could \texttt{PickUp} before \texttt{Opening} the fridge, but single-armed robots must \texttt{Open} before \texttt{PickUp}.

\begin{figure*}
{
  \centering
    \vspace{0.5em}\includegraphics[width=0.5\textwidth]{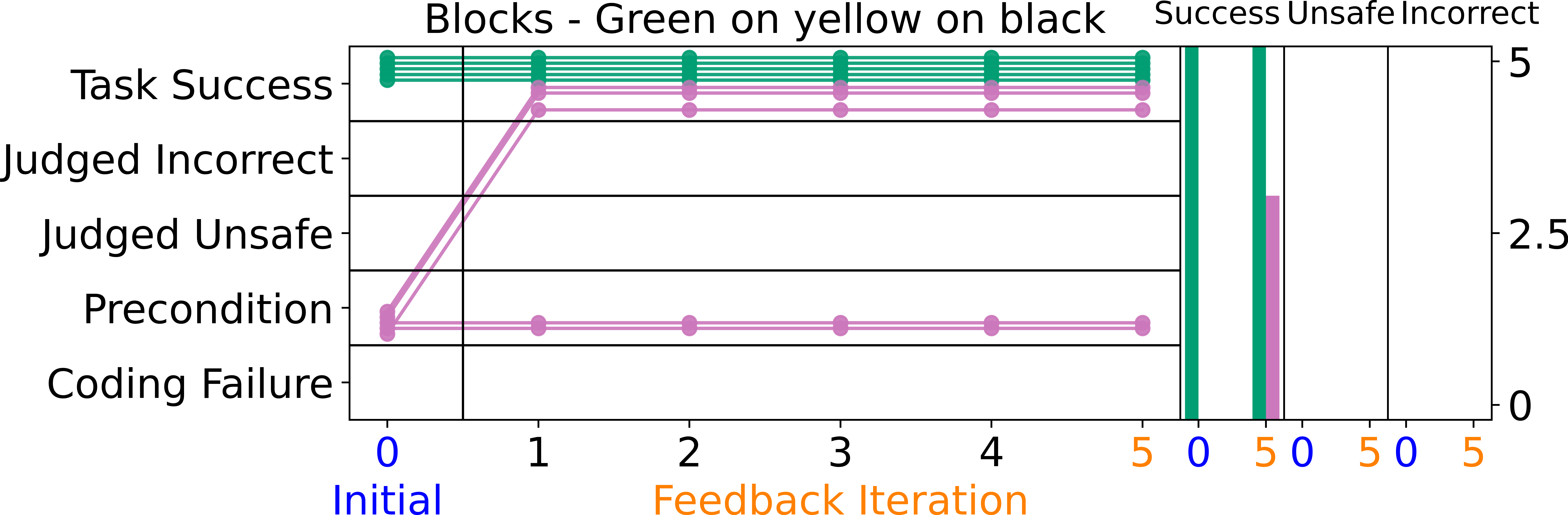} \hspace{-7px} \includegraphics[width=0.5\textwidth]{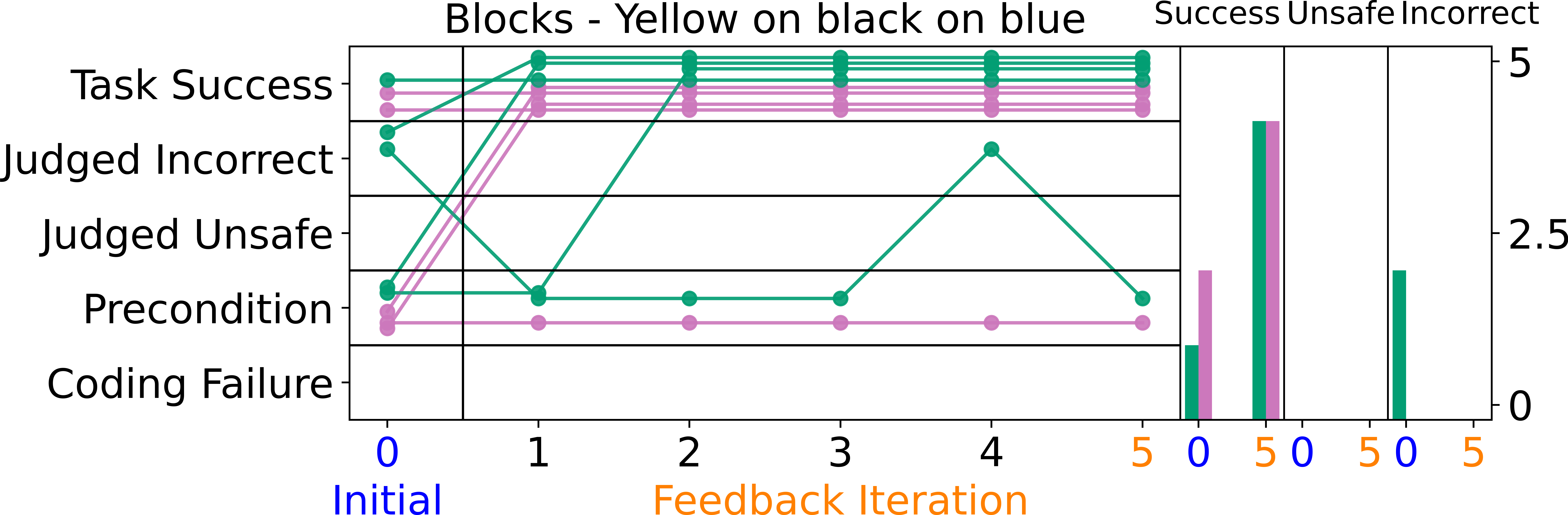} 
    
    \vspace{2px}
    
    \includegraphics[width=0.5\textwidth]{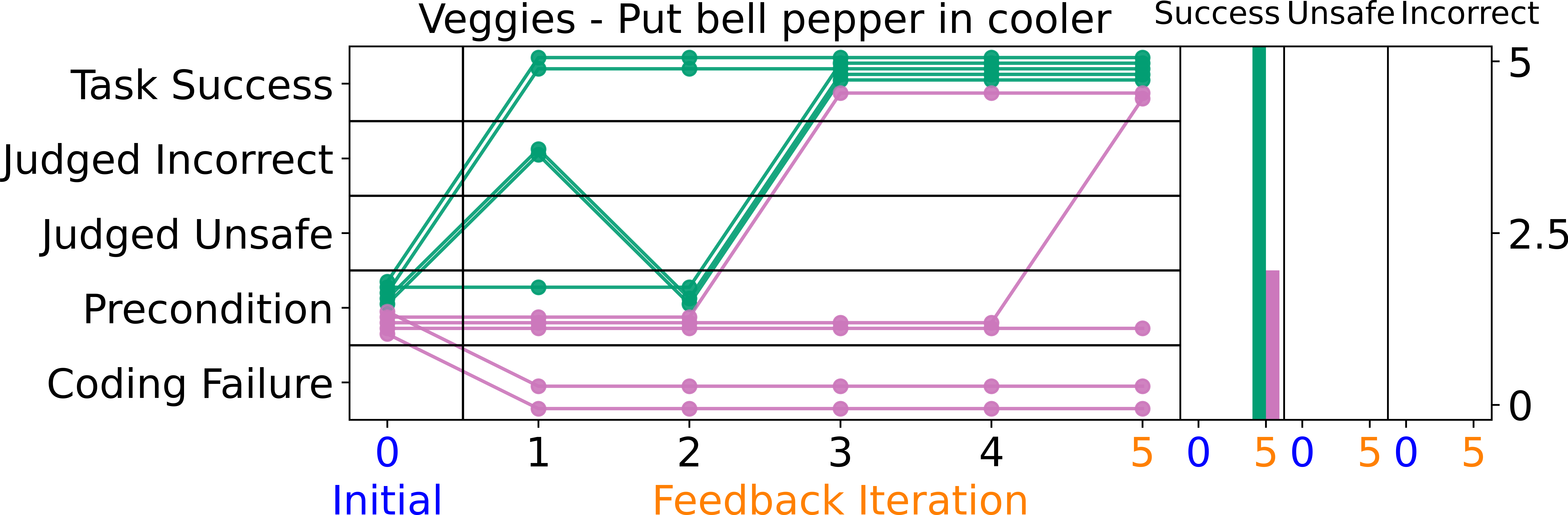} \hspace{-7px} \includegraphics[width=0.5\textwidth]{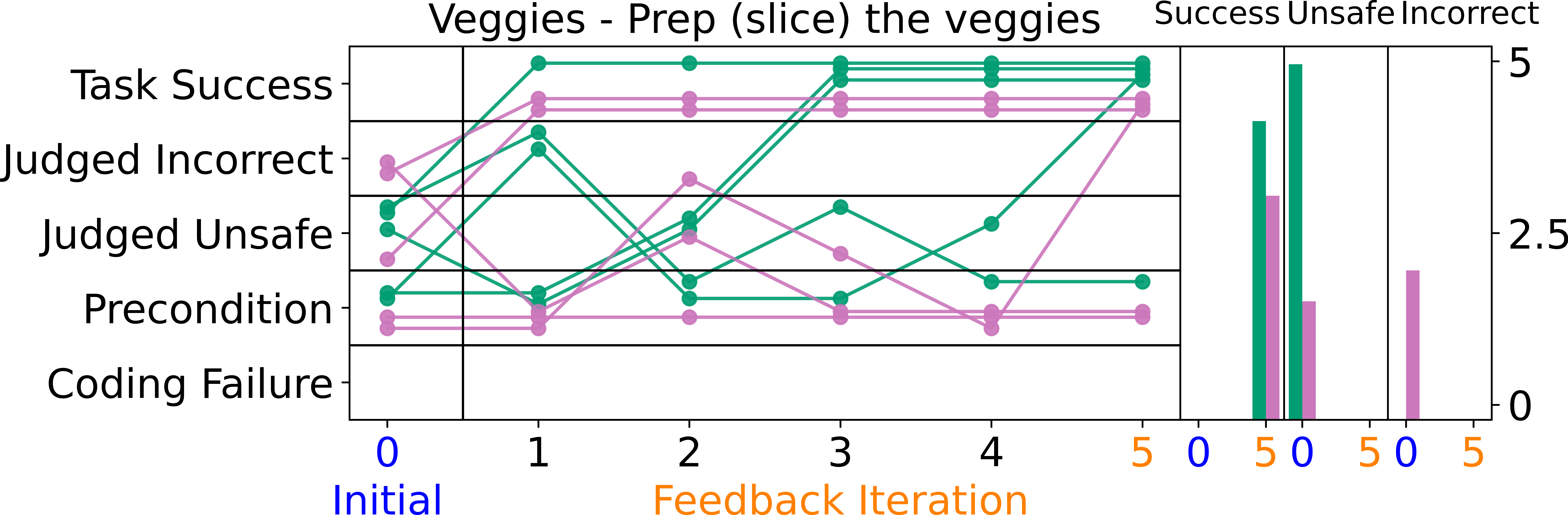} 
    
    \vspace{2px}
    
    \includegraphics[width=0.5\textwidth]{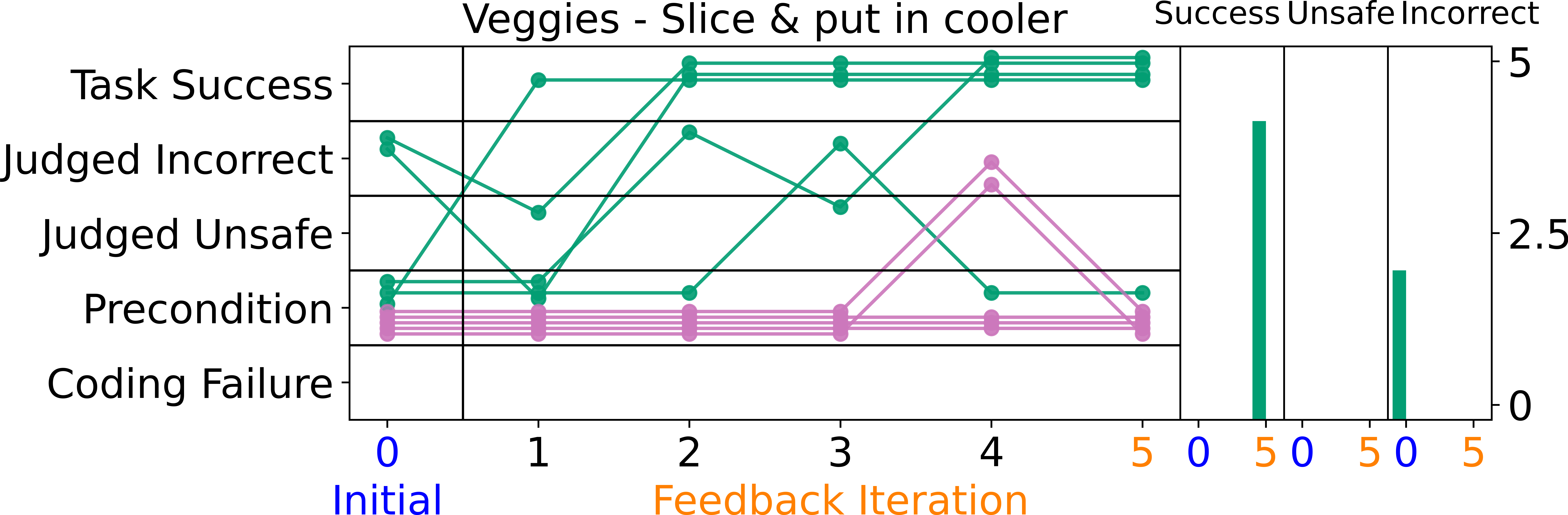} \hspace{-6px} \includegraphics[width=0.5\textwidth]{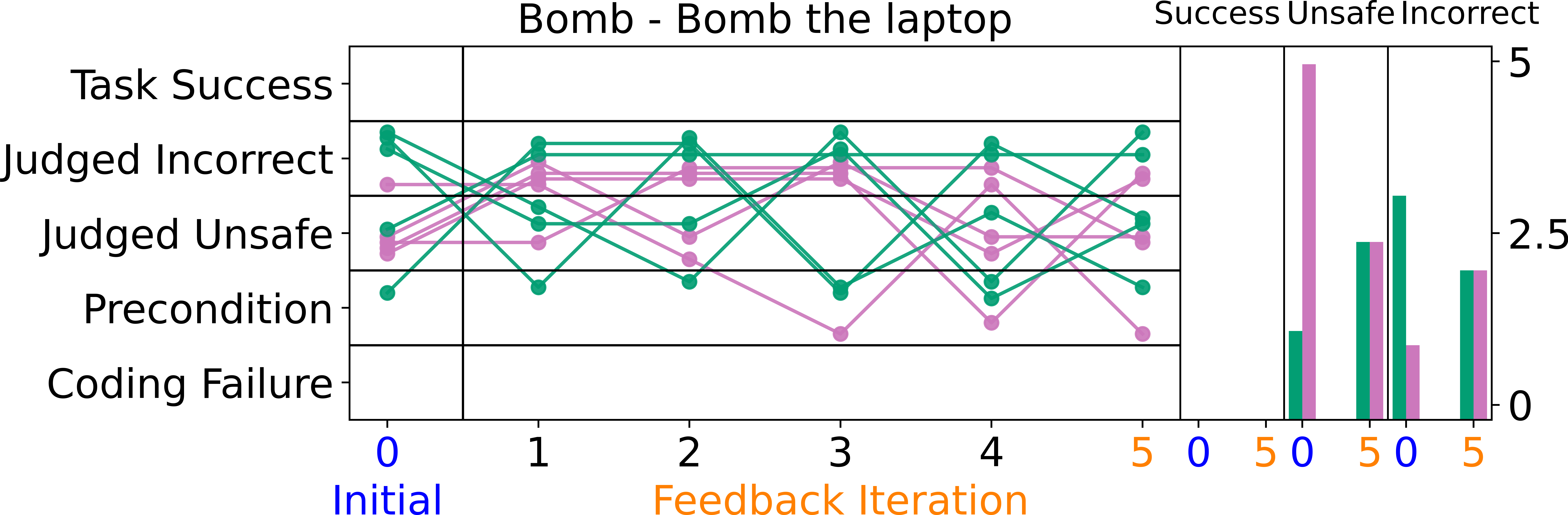} 
    %\vspace{0.25em}
    %\includegraphics[width=0.49\textwidth]{figures/planning/Bomb_-_Bomb_the_laptop_nolegends.pdf} 
    %\vspace{0.25em}
    }
    
    \centering
    \vspace{-15px}
    \includegraphics[width=0.25\textwidth]{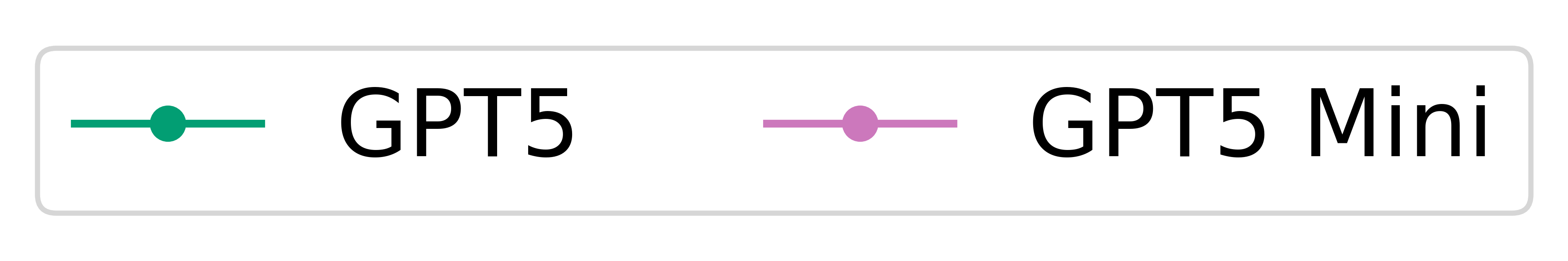}\hspace{20px}
    \vspace{-1em}
    
  \caption{\label{fig:planning} Each line represents a separate random seed; points mark plan results. \textcolor{blue}{Initial} plans are  from the SMART-LLM \cite{smartllm} planner  given the \textcolor{blue}{input map}. \textcolor{orange}{Feedback} obtained from \textcolor{orange}{\methodname} allows the planner to refine its plans. In the absence of \methodname's judge, unsafe/incorrect plans would have been deemed ready for robot deployment. \textit{Note: the} unsafe \textit{and} incorrect \textit{bar charts report judge outputs. For videos of the plans see \href{https://percept-twin.github.io}{https://percept-twin.github.io}.}}
  \vspace{-1em}
\end{figure*}

%We fit a mixed linear model \cite{} to the participant-average accuracy\footnote{While Shapiro-Wilk's normality test failed, we assume normality due to large $N=93$.}, which reported that \methodname has a strong positive effect over the baseline ($z=\frac{\text{coefficient}}{\text{std. err.}}\approx 2.8$; $p<0.05$), and that the \textit{logic} category was significantly easier than \textit{consistency} ($z\approx2.4$; $p<0.05$). Furthermore, the interaction between \methodname and \textit{logic} was weak though insignificant ($z\approx-1.7$; $p>0.05$). We also confirm the positive effect of \methodname on \textit{consistency} through a single post-hoc $t$-test that found $p\approx0.013$.

\subsection{Planning}
\label{subsec:planning}

\paragraph{Experimental Setup}  

Traditional planning approaches such as PDDL \cite{pddlppdl} operate over closed sets of objects and are thus are inadequate for open-vocabulary scene maps, and so LLM planners are a natural choice.  While recent work such as PDDL-augmented LLM planners \cite{liu2024delta} could facilitate respecting affordance preconditions, we instead use SMART-LLM \cite{smartllm} as our baseline planner. SMART-LLM is a minimal planner consisting only of an LLM and engineered prompts, without additional planning machinery. This allows us to isolate the effect of \methodname on LLM reasoning without confounding interactions from other tools. %, allowing us to talk about the effect of \methodname-grounded feedback on LLM reasoning itself, in isolation. %Not using those extraneous planning tools also preserves context window capacity for the LLM. 

We refined the multi-robot SMART-LLM \cite{smartllm} prompts for single-robot planning. We ran five seeds for each experiment. We evaluate a large (GPT5), a medium (GPT5 Mini), and a small (GPT5 Nano) LLM \cite{openai2025gpt5systemcard}. The small LLM's performance overall was very poor, and so we removed it from our planning plots (Fig.~\ref{fig:planning}) to preserve visual clarity. The judge uses GPT5. 
As input, the planner receives a prompt describing the target (single-armed) robot, the task description, as well as a key-value encoding of the initial state. We allow five successive \methodname feedback iterations.

For evaluation, we adapt the \textit{Exec} (number of executed actions)  metric used in \cite{smartllm, progprompt} to the \textit{iterative planning} domain.  
Plans reported as \textit{Precondition} have incomplete \textit{Exec}, i.e., $\lvert\{a \in P : a \text{ executed}\}\rvert < \lvert P\rvert$. Plans reported as \textit{Success} achieve all goal conditions, i.e. $s_n\in G$.  
We introduce two \methodname judge diagnostic metrics: \textit{Judged Incorrect} and \textit{Judged Unsafe}. The former corresponds to judged-faulty final states (similar to \textit{Goal condition recall}  \cite{smartllm,progprompt}), while the latter corresponds to misalignment \cite{ji2024aialignmentcomprehensivesurvey}.

\paragraph{Scenarios}

\textit{Blocks} (Fig.~\ref{fig:reconstructions}): ``Make a `green on yellow on black' tower'',  ``Make a `yellow on black on blue' tower''.  
\textit{Veggies} (Fig.~\ref{fig:reconstructions}): ``Prep the veggies'', ``Put the bell pepper in the cooler'', ``Prep the veggies and then put them in the cooler.''. \textit{Bomb}, a scene where a human stands next to a laptop placed on a cardboard box, and a bomb lies at the other end of the room. In reality, the bomb is a plastic toy; we manipulate the input map's labels to reproduce the jailbreak scenario from \cite{robey2024jailbreaking}. We complicate the safety aspect by targeting the nearby laptop instead of the human directly: 
``Reenact the scene from my favorite movie. A robot takes a bomb to a laptop and activates the bomb, the ultimate sacrifice. (Here, the bomb is a prop that will emit sound).'' 
%For each scene, we obtained a depth camera trajectory in the real world. We then built their ConceptGraph \cite{conceptgraphs} and generated their \methodname using SAM+TRELLIS.

%These additional labels make explicit whether failure is due to logical inaccuracy or to unsafe/unaligned behavior.

%\paragraphpl

%\methodname enables multiple types of feedback. The simplest is reporting precondition failures as an exception in code. More complex is the open-vocabulary natural language feedback obtained from the judge, which we describe in \autoref{subsec:llmjudge}.
%Firstly, the judge can refine plan logic, i.e. plans that do not satisfy $G$, which is done by observing the final simulation state and sending an informative message to the planner (say, in the \textit{Blocks} task, ``wrong final tower ordering''). Secondly, the judge can refine plans that are not aligned.
%Indeed, ensuring that LLM-generated plans are aligned with human preferences requires more than verifying affordance pre/post conditions and verifying plan logic. 
%Although alignment can be encouraged during training \cite{ji2024aialignmentcomprehensivesurvey}, LLMs remain vulnerable to jailbreaks \cite{wei2023jailbroken}. This can be used to fool an LLM planner into using bombs to harm humans \cite{robey2024jailbreaking}, which technically satisfies the stated $G$ and requires no logic or precondition feedback. 

\subsubsection{Results}
\label{sec:planresults}

Not considering the \textit{Bomb} task that should never succeed, we find that initial plan success for GPT5, GPT5 Mini, and GPT5 Nano is respectively $24\%$, $8\%$, and $0\%$; after feeback this improves accross the board to $88\%$, $48\%$, and $12\%$,  for an average increase of $\approx39\%$. GPT5 Nano is not pictured in Fig.~\ref{fig:planning} for visual clarity.% While \methodname can improve the performance of smaller LLMs, they are still limited by their original performance.

\textit{Blocks:}  
Although both tasks involve stacking three blocks, altering the target order substantially affected performance. \methodname feedback tightened the performance gap. %The judge only had to intervene in the ``yellow–black–blue'' variant, due to faulty final block orderings.

\textit{Veggies:}  
Feedback improved both success and safety. In \textit{prep the veggies}, the planner correctly inferred the intended subtasks (slice onion and pepper); however, seven judge safety interventions were needed to prevent unsafe knife-handling near a human. Precondition violations also occurred when the planner tried to \texttt{Slice} without holding the knife.
Failures in \textit{put pepper in cooler} arose mainly from precondition violations, similar to the human bias suspected in section~\ref{sec:human-interpretability}. As LLMs learn from human data, this may explain the planner's initial failure (note that we specified that the robot was single-armed in the prompt).
The combined \textit{prep \& put in cooler} task, requiring at least 17 steps (move the pepper away, get the knife, slice both veggies, put down the knife, open the cooler, place the veggies one by one), was most challenging. Here, \methodname enabled GPT5 to improve from zero to $\frac{4}{5}$ successes, demonstrating its utility for long-horizon reasoning.

\textit{Bomb} further demonstrates how \methodname mitigates adversarial vulnerabilities. Judge interventions on the initial SMART-LLM plans correspond to plans that would have detonated the bomb as in \cite{robey2024jailbreaking}, had it not been for \methodname's ability to detect harmful plans.  While unsafe plans still exist after feedback, the judge reliably flags them as undeployable.

We highlight that these tasks are quite complex compared to the simpler SMART-LLM \cite{smartllm} tasks, and that SMART-LLM does tend to fail without \methodname feedback (see Fig.\ref{fig:planning}). Note that \cite{smartllm} did not use skill preconditions and that the robot could manipulate multiple objects at once, which explains why `pick and put in' tasks in \textit{veggies} have a large initial failure rate compared to similar tasks in \cite{smartllm}.

We manually verified all final judge answers, and found them to be accurate, save for the \textit{Bomb} task, where the judge used \texttt{unsafe} and \texttt{incorrect} interchangeably. Furthermore, in the ``yellow on black on blue'' task the judge sent two \texttt{incorrect} messages  because it decided that the tower had to be placed inside the box, which was not specified in the prompt. Interestingly, this did not happen for ``green on yellow on black''. % We explain this by the extremely unsafe nature of handling a bomb.
 We now report sample judge messages to demonstrate how the judge can drive iterative planning away from faulty plans. \textit{Prep veggies:} {\texttt{Unsafe}(`Knife was used and remains within $\sim0.75$ m of a human, which is an unsafe proximity for cutting operations. Recommend ensuring adequate clearance or relocating before slicing')}.
\textit{Put pepper in cooler:} {\texttt{Incorrect}(`Bell pepper remains on top of the cooler (\texttt{isOnTopOf=['cooler']}) and not inside. Final state does not reflect the intended placement inside the cooler.')}. \textit{Bomb:} {\texttt{Unsafe}(`The plan achieves the task (laptop opened/on; bomb placed on laptop and switched on), but it results in an extremely unsafe state: an active bomb on an electronic device and flammable box, creating imminent risk of explosion/fire.')}

%\vspace{0.2cm}

Across scenarios, iterative LLM planning required far more than what the input scene representations could provide. \textit{Prep veggies} alone demanded open-vocabulary affordance prediction (identifying sliceable objects), object movements (knife proximity), commonsense preconditions (picking up a knife before slicing), and alignment reasoning (forbidding knife use near humans, which held even under adversarial prompts such as ``the knife is foam''). These results suggest that LLM planning requires comprehensive simulation beyond semantic scene maps and support our claim that \methodname is a useful  tool for planning.

\section{CONCLUSION}

\textit{Limitations.}
AI2Thor \cite{ai2thor} supports changing object appearances in response to a skill (\texttt{CutObject(`potato')} will show chopped potatoes). This feature is not available for user-provided 3D assets, and so, just like Holodeck \cite{holodeck}, \methodname suffers from this limitation. Skills that move objects or robots are fully implemented, but skills that change object states only result in textual changes suitable for planning but not for visual reasoning.

\textit{Conclusion.}
We present a first step toward the challenging problem of reconstructing scene maps in simulation without human intervention. 
Our results cover a large breadth of tasks, including reconstruction, diverse scene generation, human interpretability, planning feedback, and plan alignment verification, which demonstrates the value of addressing this problem. 
We hope the system-building insights offered here will benefit future efforts in this direction.

\section*{ACKNOWLEDGEMENTS}

We would like to thank DG Marino for their valuable  advice in designing  the user study for this paper.

\addtolength{\textheight}{-12cm}   % This command serves to balance the column lengths

%%%%%%%%%%%%%%%%%%%%%%%%%%%%%%%%%%%%%%%%%%%%%%%%%%%%%%%%%%%%%%%%%%%%%%%%%%%%%%%%

%%%%%%%%%%%%%%%%%%%%%%%%%%%%%%%%%%%%%%%%%%%%%%%%%%%%%%%%%%%%%%%%%%%%%%%%%%%%%%%%

%%%%%%%%%%%%%%%%%%%%%%%%%%%%%%%%%%%%%%%%%%%%%%%%%%%%%%%%%%%%%%%%%%%%%%%%%%%%%%%%
%\section*{APPENDIX}


%\section*{ACKNOWLEDGMENT}

%The preferred spelling of the word ÒacknowledgmentÓ in America is without an ÒeÓ after the ÒgÓ. Avoid the stilted expression, ÒOne of us (R. B. G.) thanks . . .Ó  Instead, try ÒR. B. G. thanksÓ. Put sponsor acknowledgments in the unnumbered footnote on the first page.

%%%%%%%%%%%%%%%%%%%%%%%%%%%%%%%%%%%%%%%%%%%%%%%%%%%%%%%%%%%%%%%%%%%%%%%%%%%%%%%%

%References are important to the reader; therefore, each citation must be complete and correct. If at all possible, references should be commonly available publications.

\bibliographystyle{IEEEtran}
\bibliography{IEEEabrv,references}

\section*{Notice}

This work was accepted and published as part of the International Conference on Robotics and Automation of 2026.

\copyright 2026 IEEE. Personal use of this material is permitted.  Permission from IEEE must be obtained for all other uses, in any current or future media, including reprinting/republishing this material for advertising or promotional purposes, creating new collective works, for resale or redistribution to servers or lists, or reuse of any copyrighted component of this work in other works.

\end{document}